\documentclass[10pt,twocolumn,letterpaper]{article}

\usepackage[applications]{wacv} 
\usepackage{xspace} 
\usepackage{graphicx}
\usepackage{array}
\usepackage{tabularx}
\usepackage{booktabs}
\usepackage[export]{adjustbox} 

\newcommand{\placeholderheight}{38mm}

\newcolumntype{M}[1]{>{\raggedright\arraybackslash}m{#1}}
\newcolumntype{C}[1]{>{\centering\arraybackslash}m{#1}}
\newcolumntype{Y}{>{\centering\arraybackslash}X}

\usepackage{subcaption}
\newcommand{\formimg}[2][]{%
  \fbox{
    \begin{minipage}[c][\placeholderheight][c]{\linewidth}
      \centering
      \includegraphics[width=\linewidth,height=\placeholderheight,#1]{#2}
    \end{minipage}%
  }%
}

\newcommand{\pageimg}[2][]{%
  \fbox{
    \begin{minipage}[c][48mm][c]{\linewidth}
      \centering
      \includegraphics[width=\linewidth,keepaspectratio,height=48mm,#1]{#2}
    \end{minipage}%
  }%
}

\DeclareRobustCommand{\commonforms}{\textsc{CommonForms}\xspace}

\definecolor{wacvblue}{rgb}{0.21,0.49,0.74}
\usepackage[pagebackref,breaklinks,colorlinks,allcolors=wacvblue]{hyperref}

\def\wacvPaperID{2211} 
\def\confName{WACV}
\def\confYear{2026}

\title{\commonforms: A Large, Diverse Dataset for Form Field Detection}

\author{Joe Barrow\\
Independent Researcher\\
{\tt\small joseph.d.barrow@gmail.com}
}

\begin{document}
\maketitle
\begin{abstract}
This paper introduces CommonForms, a web-scale dataset for form field detection. 
It casts the problem of form field detection as object detection: given an image of a page, predict the location and type (\texttt{Text Input}, \texttt{Choice Button}, \texttt{Signature}) of form fields.
The dataset is constructed by filtering Common Crawl to find PDFs that have fillable elements.
Starting with 8 million documents, the filtering process is used to arrive at a final dataset of roughly 55k documents that have more than 450k pages.
Analysis shows that the dataset contains a diverse mixture of languages and domains; one third of the pages are non-English, and among the 14 classified domains, no domain makes up more than 25\% of the dataset.

In addition, this paper presents a family of form field detectors, FFDNet-Small and FFDNet-Large, which attain a very high average precision on the CommonForms test set.
Each model cost less than \$500 to train.
Ablation results show that high-resolution inputs are crucial for high-quality form field detection, and that the cleaning process improves data efficiency over using all PDFs that have fillable fields in Common Crawl.
A qualitative analysis shows that they outperform a popular and commercially available PDF reader that can prepare forms.
Unlike the most popular commercially available solutions, FFDNet can predict checkboxes in addition to text and signature fields.
This is, to our knowledge, the first large-scale dataset released for form field detection, as well as the first open-source models. 
The dataset, models, and code will be released at \url{https://github.com/jbarrow/commonforms}.

\end{abstract}
    
\begin{figure*}[!ht]
  \centering
  \includegraphics[width=\textwidth]{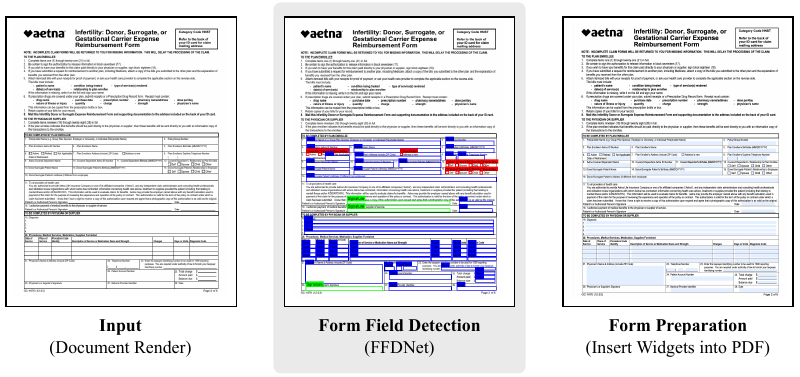}
  \caption{Overview of form field detection and preparation.}
  \label{fig:overview}
\end{figure*}

\section{Introduction}
\label{introduction}
Despite decades of digitalization, a large volume of real-world transactions still center around paper forms: insurance claims, municipal paperwork, school permission slips, and many, many others.
These documents are often distributed as scans or non-fillable (``flat'') PDFs, and require either printing, tedious software workarounds, or turning to a proprietary solution like Adobe Acrobat or Apple Preview.
In order to be digitally filled, a flat form must be prepared by having interactive form fields inserted.
Although proprietary solutions may employ machine learning to prepare forms, there is not yet a high-quality open-source machine learning-based system that can automatically and reliably prepare fillable forms.

Converting a flat PDF to an \emph{accessible} form --- one that can be understood by a screen reader, or filled automatically --- typically requires two steps:
\begin{enumerate}
\itemsep0em
\item \textbf{form field detection}: detecting the locations and types (e.g. text or checkbox) of fillable elements; and
\item \textbf{form enrichment}: grouping the fillable elements and their labels based on the semantics of the form.
\end{enumerate}

Previous work has primarily focused on the second~\citep{aggarwal-etal-2020-form2seq,aggarwal2020multi}.
In this work, we tackle the first problem by detecting where the fillable elements should go visually in a PDF.
To this end, we construct and release both a large-scale dataset, \commonforms, consisting of more than 480k pages from more than 59k document forms.

We also train and release a family of form field detection models trained on this dataset, FFDNet-Small and FFDNet-Large.
Training each of these models \textbf{costs roughly \$500 of compute or less}, and yet these models are capable of high-quality form field detection.
They achieve a high average precision of more than 80 on the \commonforms test set.
We perform a qualitative analysis of these models against Adobe Acrobat and show that FFDNet has better recall and precision than Acrobat.

\commonforms is drawn from a large collection of interactive PDF forms scraped from the Internet.
The core insight of this paper is that ``quantity has a quality all its own,'' and that we can leverage existing fillable forms as a training signal.
We use Common Crawl as a wellspring of PDFs and apply a rigorous cleaning process.
This cleaning process results in improved data efficiency compared to using every PDF with a form field.

To train the FFDNet family of models, we cast the problem of form field detection as a pure object detection problem.
Given a page image, the goal is to predict the location and types of each form element.
The types are drawn from teh following set: \{\texttt{Choice Button}, \texttt{Text Input}, \texttt{Signature}\}.
This means that, unlike Adobe Acrobat and Apple Preview, FFDNet can predict the location of choice buttons (i.e. checkboxes and radio buttons).
As of the time of publication, these tools only predict text form fields.

The FFDNet models are high-resolution (1216px) YOLO11~\citep{yolo11_ultralytics} object detectors.
Our ablation results show that high-resolution inputs are crucial to quality; models trained on 10k pages at varying resolutions show a range of roughly 20 mean average precision points.

Our contributions are as follows:
\begin{itemize}
\item we prepare \textbf{release \commonforms}, a large, diverse, and high quality dataset for form field detection;
\item we train and \textbf{release a family of form field detection models} on \commonforms, FFDNet-Small and FFDNet-Large\footnote{supported by a LambdaLabs compute grant};
\item we provide an \textbf{in depth analysis of \commonforms} to identify the represented languages and domains; and
\item we \textbf{compare the FFDNet models against the most popular commercial system}, Adobe Acrobat, and show that they produce substantially higher quality forms.
\end{itemize}
\section{Related Work}
\label{sec:related_work}

Machine-learning-based work on document forms can be viewed as two broad categories: form preparation and form understanding.
Form preparation is the task of making a form fillable, whether that is detecting widgets or building a semantic model of the form.
Form understanding, also known as key information extraction (KIE), is an information extraction task on filled-out document forms.
This paper is an instance of form preparation, which is the smaller of the two subfields.

\paragraph{Form Preparation and Understanding} 
Representative work on form preparation includes the line of work on Form2Seq~\citep{aggarwal-etal-2020-form2seq,aggarwal2020multi} and FormA11y~\citep{paliwal2025forma11y}.
Form2Seq formalizes a semantic form model, treating it as a joint classification and grouping task.
Lower-level elements like widgets and text blocks are first classified into, e.g. \texttt{ChoiceButton} or \texttt{TextField}, and then grouped into higher-order categories like \texttt{ChoiceGroup}.
However, it is assumed that the interactive form widgets have already been correctly detected.
FormA11y takes a human-in-the-loop approach in which users match labels to widgets to create accessible forms.
In contrast to both of these, \commonforms introduces a large-scale dataset and baseline models for automatic detection of the form fields themselves, allowing the end-to-end  preparation of a form from a flat document.

Form understanding (KIE) focuses on extracting key-value pairs from completed forms.
Common form understanding datasets tend to be small, such as FUNSD~\citep{jaume2019funsd} (199 labeled forms), and NAF~\citep{davis2019deep} (77 labeled forms). Methodologically, work in the area has employed a diverse body of models, including: purely visual models, such as image segmentation~\citep{vu2020revising}; purely textual models, such as BERT~\citep{devlin2019bert}; multimodal models that can model vision, text, and layout information jointly, such as the LayoutLM series~\citep{xu2020layoutlm,xu2020layoutlmv2,huang2022layoutlmv3}; as well as vision and graph-based models such as Visual FUDGE~\citep{davis2021visual}. 
In this work, we tackle form \emph{preparation} rather than understanding, and cast the task as one of object detection.
That is, we do not attempt to extract values, but to predict where on the page the slots for values should be, based on visual information.

\paragraph{Object Detection in Document Images}
Many documents are inherently multimodal, so there are many established lines of research that cast document problems as vision problems, and specifically object detection.
Examples include layout detection (DocLayNet~\citep{pfitzmann2022doclaynet}, PubLayNet~\citep{zhong2019publaynet}, Newspaper Navigator~\citep{lee2020newspaper}, PRImA Layout~\citep{antonacopoulos2009realistic} ), table detection (TableBank~\citep{li2020tablebank}), and math formula detection (FormulaNet~\citep{schmitt2022formulanet}).
LayoutParser~\citep{shen2021layoutparser} is a suite of tools that provides a unified interface for all of these tasks.
Depending on the resolution required, models such as YOLO~\citep{yolo11_ultralytics}, the Document Image Transformer (DIT)~\citep{li2022dit}, or LayoutLM~\citep{xu2020layoutlm}, for multimodality, have been used.
\commonforms adopts the same paradigm for detecting form fields.
In this work, we treat documents as unimodal images and train object detectors to localize and type form fields.
The outputs can be used as a complementary semantic layer in addition to these other models.

\paragraph{Document Corpora from Common Crawl}
ccPDF~\citep{turski2023ccpdf} and FinePDFs~\citep{kydlicek2025finepdfs} curate PDF corpora from Common Crawl.
Both datasets target visually and/or topically diverse datasets.
Like these efforts, \commonforms is also drawn from Common Crawl.
However, in this work, the filtering and preparation is focused on mining well-annotated forms which can be used as training signal for form field detection.
\begin{figure}[!ht]
  \centering
  \includegraphics[width=0.48\textwidth]{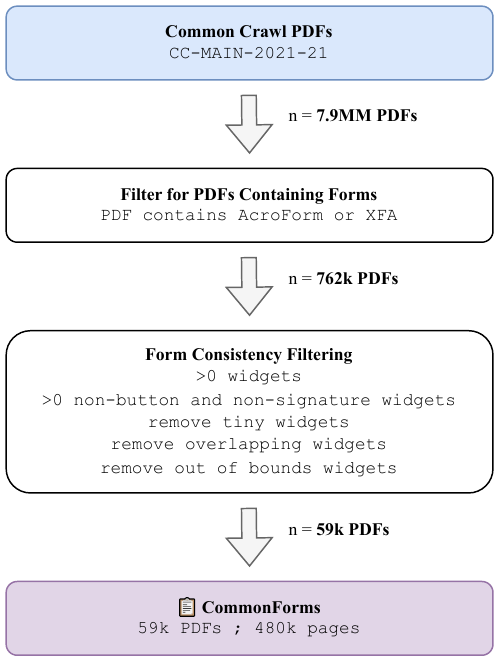}
  \caption{Filtering pipeline for \commonforms.}
  \label{fig:filtering}
\end{figure}

\begin{table*}[p]
\centering
\small
\caption{Common inconsistencies in real-world PDF forms. This is not an exhaustive list, merely a representative list.}
\label{tab:form-inconsistencies}
\setlength{\tabcolsep}{5pt}
\renewcommand{\arraystretch}{1.2}
\begin{tabularx}{\textwidth}{M{0.255\textwidth} C{0.34\textwidth} C{0.34\textwidth}}
\toprule
\textbf{Form Inconsistency} & \textbf{Example 1} & \textbf{Example 2} \\
\midrule

{\textbf{For Official Use Only}\par\footnotesize Sections marked ``For Official Use'' or similar are sometimes made electronically fillable, and sometimes they are left unfillable.}
& \formimg{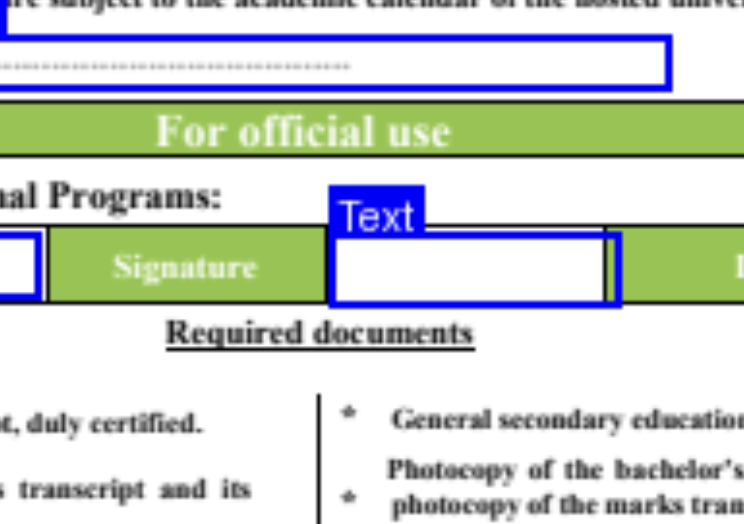}
& \formimg{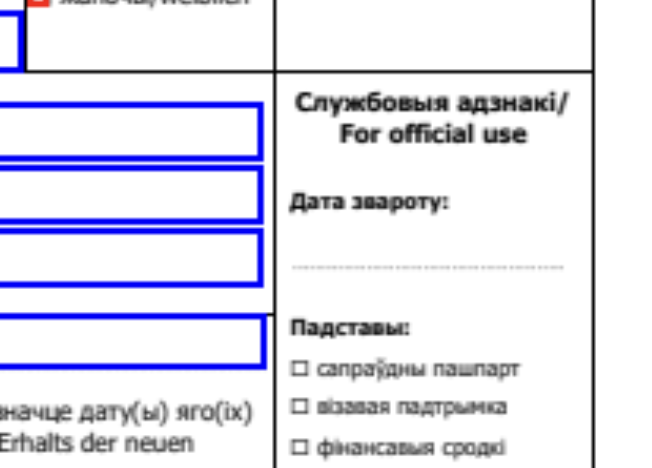} \\
\midrule

{\textbf{Circle/Check All that Apply}\par\footnotesize Print-oriented fields are digitally ambiguous. In many instances they are lef unfillable, while in others they are annotated with choice buttons.}
&
\formimg{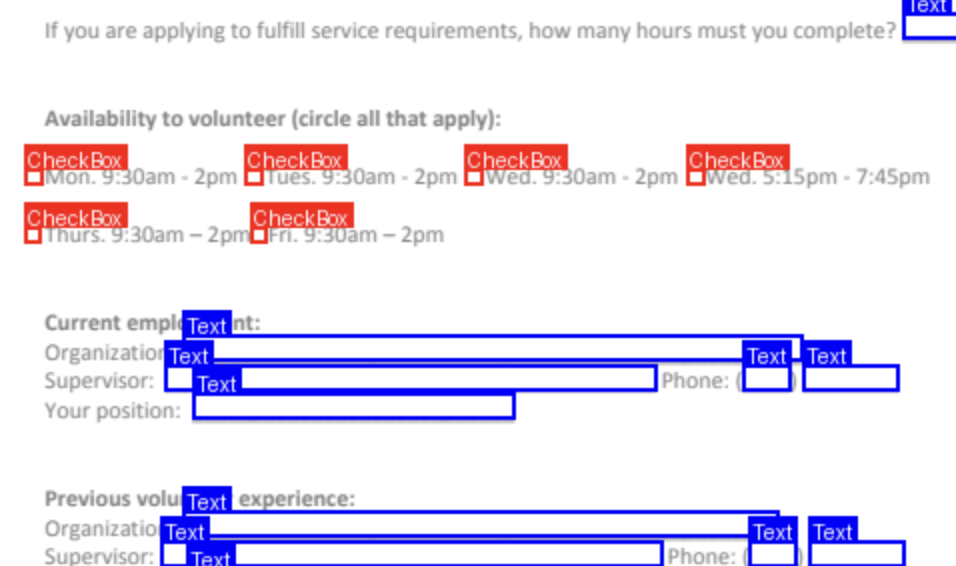}
& 
\formimg{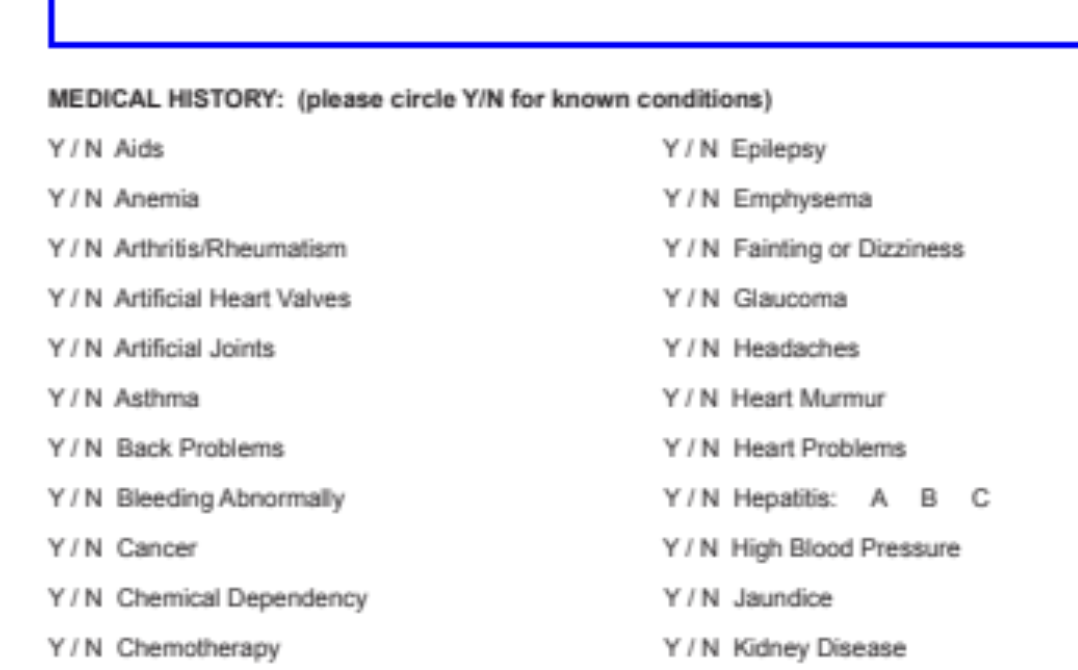}
\\
\midrule

{\textbf{Signatures as Text Fields}\par\footnotesize Signature areas are sometimes left blank, sometimes left as PDF \texttt{Signature} widgets, and sometimes implemented as PDF \texttt{Text} widgets.}
& 
\formimg{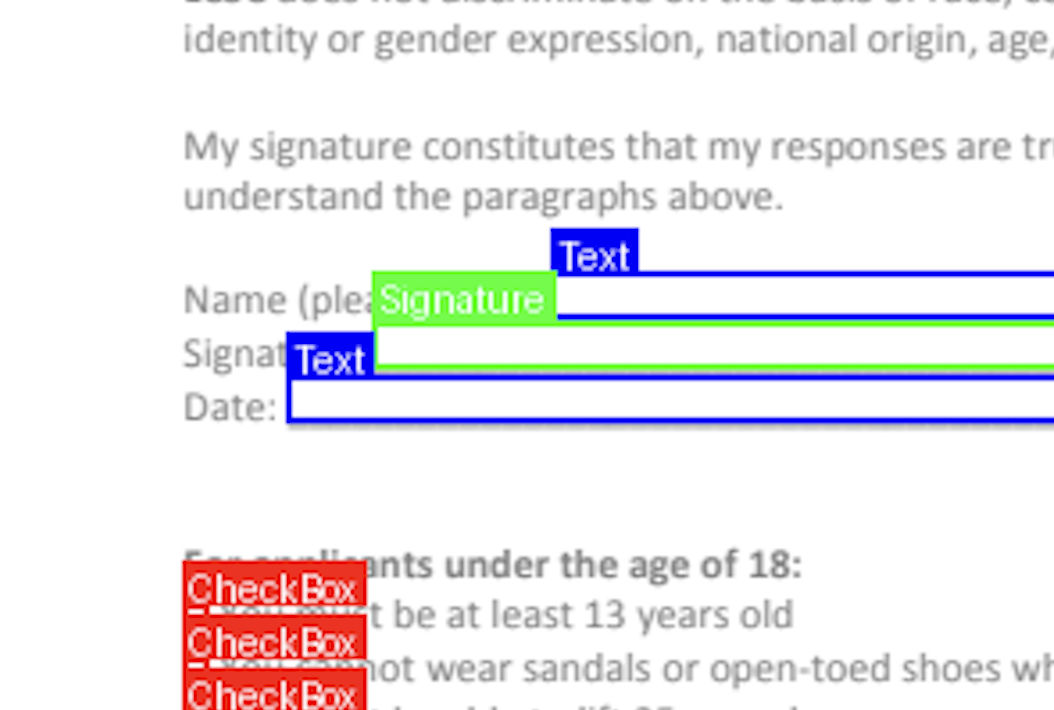}
& 
\formimg{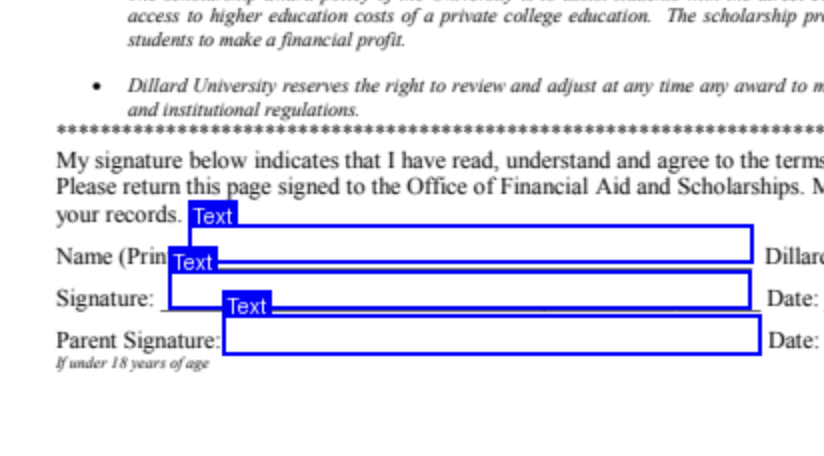}
\\
\midrule
{\textbf{Scans}\par\footnotesize Rasterized scans are often not fillable, even in the middle of a fillable PDF. They can be rotated, deformed, and noisy compared to a born-digital PDF.}
&
\formimg{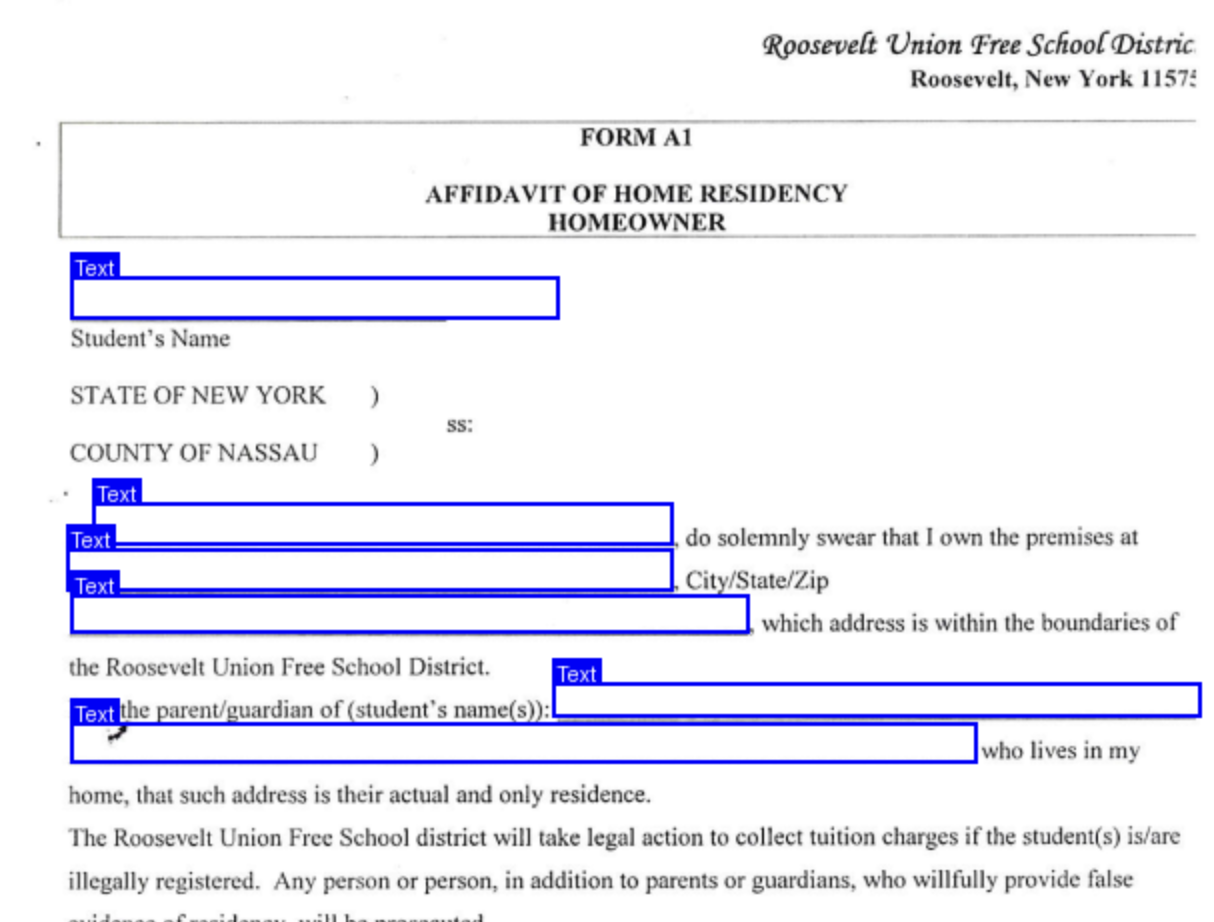}
&
\formimg{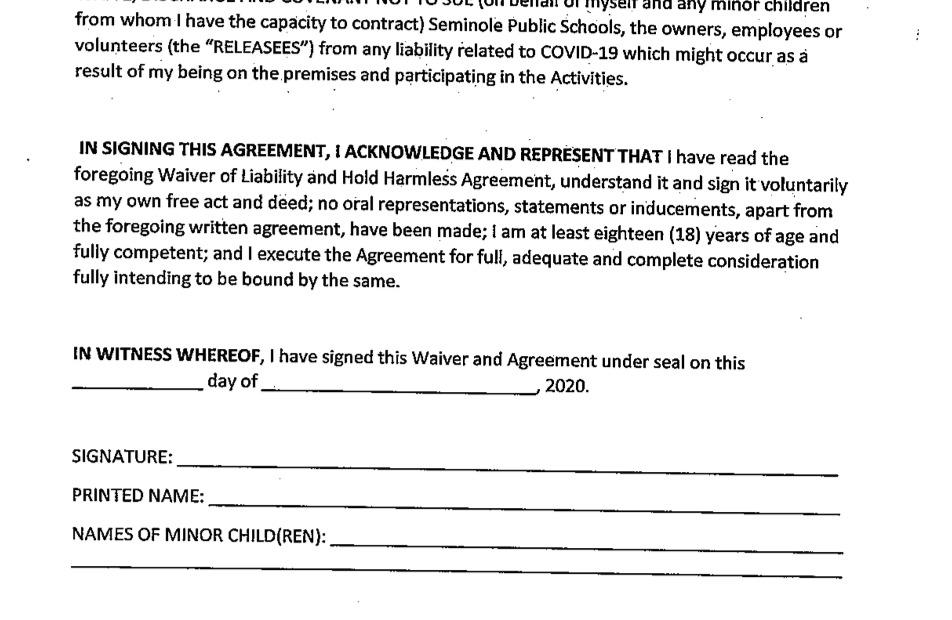}
\\ \midrule
{\textbf{Incorrect Use of Form Fields}\par\footnotesize Fields placed in repeating headers/footers, randomly placed fields in documents, or forms that have been noisily prepared.}
& \formimg{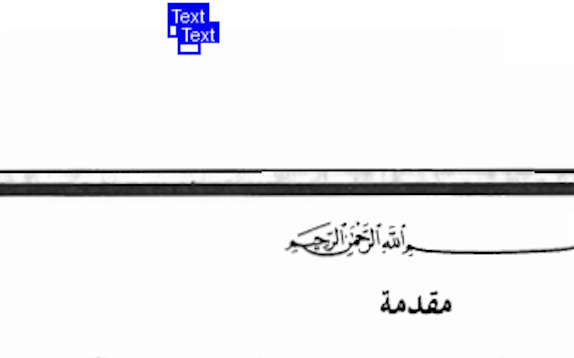}
& 
\formimg{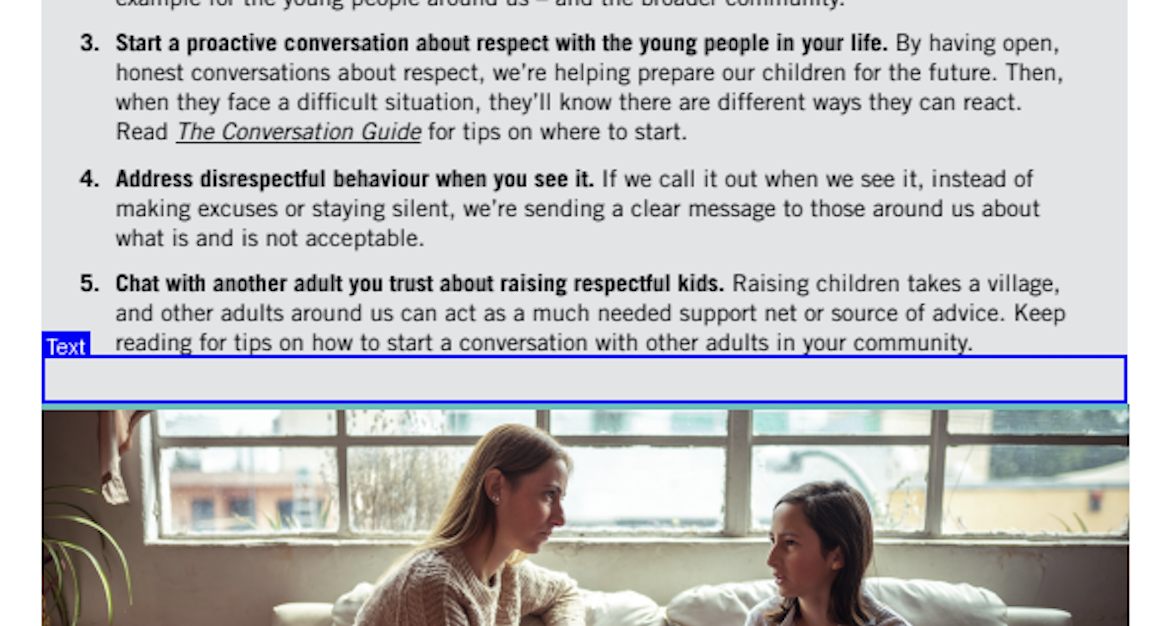}
\\
\bottomrule
\end{tabularx}
\end{table*}

\begin{figure*}[!ht]
  \centering
  \begin{subfigure}[t]{0.75\textwidth}
    \centering
    \includegraphics[width=\textwidth]{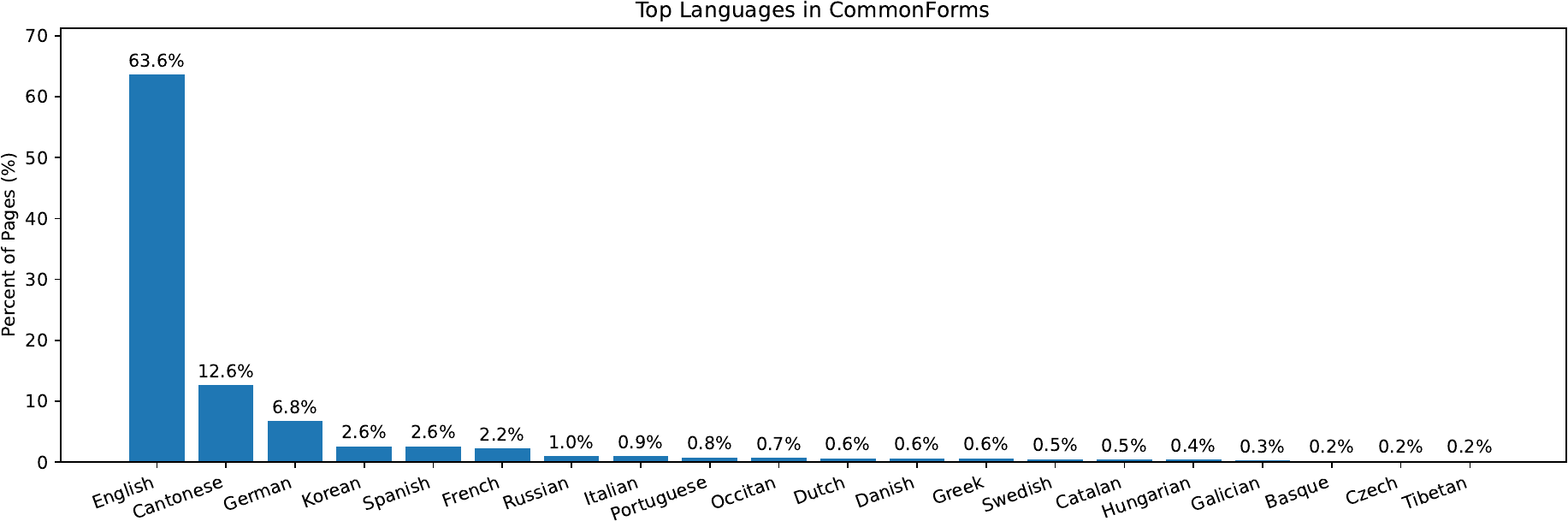}
    \caption{Language distribution of the top 20 languages. One third of the documents are non-English.}
    \label{fig:lang-share}
  \end{subfigure}

  \begin{subfigure}[t]{0.76\textwidth}
    \centering
    \includegraphics[width=\textwidth]{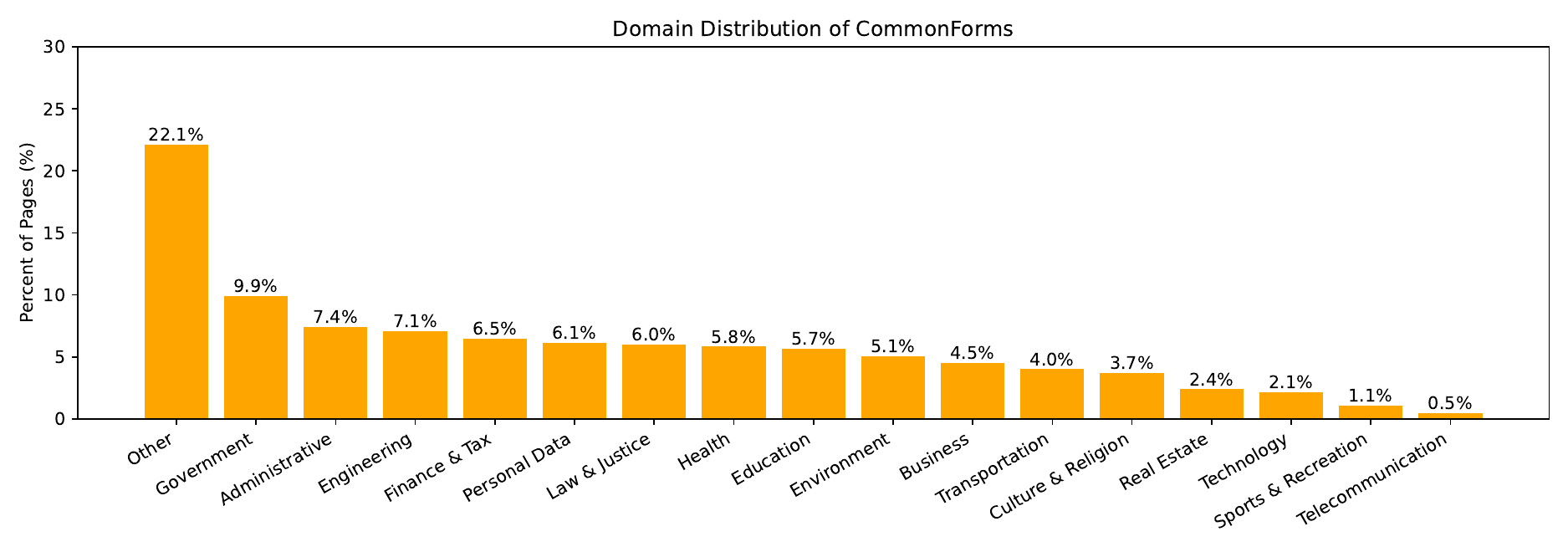}
    \caption{Domain distribution of 14 domains.}
    \label{fig:domain-hist}
  \end{subfigure}

  \caption{Distributions showing the diverse set of languages and domains represented in \commonforms.}
  \label{fig:lang-domain}
\end{figure*} 

\section{\commonforms Dataset}
\label{sec:dataset}

There is truth to the adage ``quantity has a quality all its own''.
The core thesis of this paper is twofold: (1) there are plentiful existing prepared forms in Common Crawl, and (2) those forms are high-quality enough to be used as a training signal.
With a few simple filters, we can bootstrap an effective detector without the need for manual annotation.
Our goal is to filter to forms that have been well-prepared.
We start with a \~8 million PDF sample of Common Crawl prepared by the PDF Association~\citep{wyatt2023pdfcorpus} and apply a rigorous cleaning process to arrive at the \commonforms dataset.
This filtering process is shown in Figure~\ref{fig:filtering}, and the limitations of this filtering process are discussed in Section~\ref{sec:inconsistencies}.

\subsection{Dataset Preparation and Cleaning}
In order to make use of forms from the Web, we build a filtering pipeline that starts with our candidate set and gradually reduces them to a clean set of forms.
At every stage, the pipeline applies a set of filter functions, shown in Figure~\ref{fig:filtering}.
We start with 7.9 million PDFs drawn from Common Crawl gathered from the July/August 2021 scrape~\citep{wyatt2023pdfcorpus}.
The first set of filtering functions is used to find PDFs that contain form objects.

There are two standards for PDF forms: AcroForms and the deprecated XML Forms Architecture (XFA)~\citep{ISO32000-1-2008}.
We filter only documents that contain form objects from either of these standards, reducing the pool of documents by about 90\%, to ~762k PDFs.

A PDF having a form \emph{object} does not mean that it has a well-annotated form, or even a form at all.
The next set of filtering functions is used to improve the likelihood that a document in the dataset is well annotated.
We remove documents that contain no form fields or that contain only \texttt{Button} form fields.
This second round of filtering reduces the set of forms by >90\% once again, resulting in 59k PDFs.

To improve annotations, we clean the form fields themselves, removing ones that are marked as outside the box of the page, that are too small to resolve, or that have high enough overlap with existing elements as to be considered near duplicates.
In total, \commonforms is 480k PDF pages.

We split the data into a training, validation, and test set.
We split the train set by document, rather than page, to ensure that similar pages used in training do not leak into the validation and test sets.
We build an 8k page validation set and a 25k page test set, reserving the rest of the documents for training.

\subsection{Annotation Consistency}
\label{sec:inconsistencies}

Forms from the Web are not consistently annotated.
Despite extensive filtering that reduces the candidate set of documents by more than 99\%, there are still annotation inconsistencies in the prepared forms.
These can negatively impact the real-world performance of any models trained on \commonforms.
We provide a representative, but not exhaustive, catalog of such inconsistencies in Table~\ref{tab:form-inconsistencies}.

Some of these arise from unconventional or incorrect use of form elements.
A common pattern is to see form fields used as headers and footers in a document.
The automatically prepared forms can suffer from misleading heuristics, which often look for straight horizontal lines where form fields would likely be placed.
This leads to the inclusion of spurious form fields.
Text fields can be used in place of signature fields, or signature fields are left blank, intended for a wet signature rather than a digital signature.

However, some of these are related to the semantics of the form itself.
``For Official Use Only'' sections are sometimes fillable and sometimes not.
Similarly, forms with ``Circle All that Apply'' sections are only occasionally interactive, even if the rest of the form is interactive.

Qualitative results show that despite these inconsistencies, models trained on \commonforms are eminently useful, even on complex forms.
These systematic inconsistencies provide an adverse training signal, and are a consequence of using scraped forms rather than manually annotating a dataset.
However, they are a trade-off made to scale the dataset to a practical size.
We do not attempt to filter any of these annotation inconsistencies and leave that effort for future work.

\subsection{Language Identification}

Although many cues for where a form field should go are visual, such as an underline or empty box, there are many textual cues as well.
Cues such as colons before blank spaces, circle or check all that apply sequences, and column headers all use text to indicate the presence of a form field.
There is also a difference in form field placement in right-to-left and left-to-right language.
As such, models trained to detect form fields benefit from a large number of examples for each language.

We perform language identification on \commonforms by extracting all of the text from every page, and FastText~\citep{joulin2016bag} to classify the likeliest language per page.
We show a breakdown of the top 10 most common languages and our form field detection results on them in Table~\ref{tab:subcategory-ffnet}.

Unsurprisingly, English-language forms make up the majority of the dataset at 63.6\%.
However, the other third of forms come from a broad set of languages and language families, including Cantonese, German, Korean, Spanish, French, and others.
As part of \commonforms, we release the per-page text, the language, and the classified domains.

\subsection{Domain Classification}
In order to understand the domains from which the forms originate, we use topic modeling~\citep{blei2003latent}, both to identify candidate domains and to classify the pages.
To train topic models, we repurpose the extracted text from all pages, remove stopwords, and train a topic model using LDA with MALLET~\citep{mccallum2002mallet}.
The topic model was trained with 300 topics.
We then used GPT-5 to label each topic with a primary domain and a language.
Mixed or unclear topics were labeled as Other, and English-language topics were manually verified.
The prompt, code, and topic state are all released as artefacts alongside the data.

This process resulted in 14 domains.
The results of the domain classification are shown in Table~\ref{tab:subcategory-ffnet}.
Outside of Other, the five most common domains were (1) Government and Administrative, (2) Commerce and Tax, (3) Engineering, (4) Data and Privacy, and (5) Law and Justice.
\section{FFDNet}
\label{sec:ffdnet}

We cast form field detection as an object detection task with three classes (the 4 widget types available in a PDF):
\texttt{Choice Button}, which encompasses checkboxes and radio buttons; \texttt{Text Input}; and \texttt{Signature}.
We train and release two object detectors based on YOLO11, initialized from scratch: \textbf{FFDNet-Small, (9 million parameters) and FFDNet-Large, (25 million parameters)}\footnote{Due to a processing error, both models are only trained on \~350k form pages rather than the full 490k. We are working on retraining the models for release.}.

Both FFDNet models are trained to accept high-resolution inputs, at 1216px.
This is substantially higher resolution than traditional object detection tasks, but later experiments and results in Table~\ref{tab:resolution} support the necessity of high-resolution models.
We use 1216px as it balances computational efficiency of training and inference against performance; as resolution scales beyond this batch sizes decrease and training times stretch.

The models are trained for 300 epochs with an initial learning rate of 0.001.
They are both trained on 4xV100 instances\footnote{Generously supplied via compute grant from LambdaLabs.}.
FFDNet-L took roughly 5 days to train, and FFDNet-S took roughly 2 days to train.
Hyperparameters and input resolution were chosen by training several models on subsets of \commonforms and seeing how they generalize as model and dataset sizes scale.
\begin{table*}[!ht]
\centering
\small
\caption{Qualitative comparison between Adobe Acrobat and FFDNet-S/L. Acrobat does not predict checkboxes, and has substantially lower precision and recall for text and signature form fields than FFDNet.}
\label{tab:qualitative-comparison}
\setlength{\tabcolsep}{4pt}
\renewcommand{\arraystretch}{1.1}
\begin{tabularx}{\textwidth}{YYYY}
\toprule
\textbf{Input} & \textbf{Adobe Acrobat} & \textbf{FFDNet-S (ours)} & \textbf{FFDNet-L (ours)} \\
\midrule

\pageimg{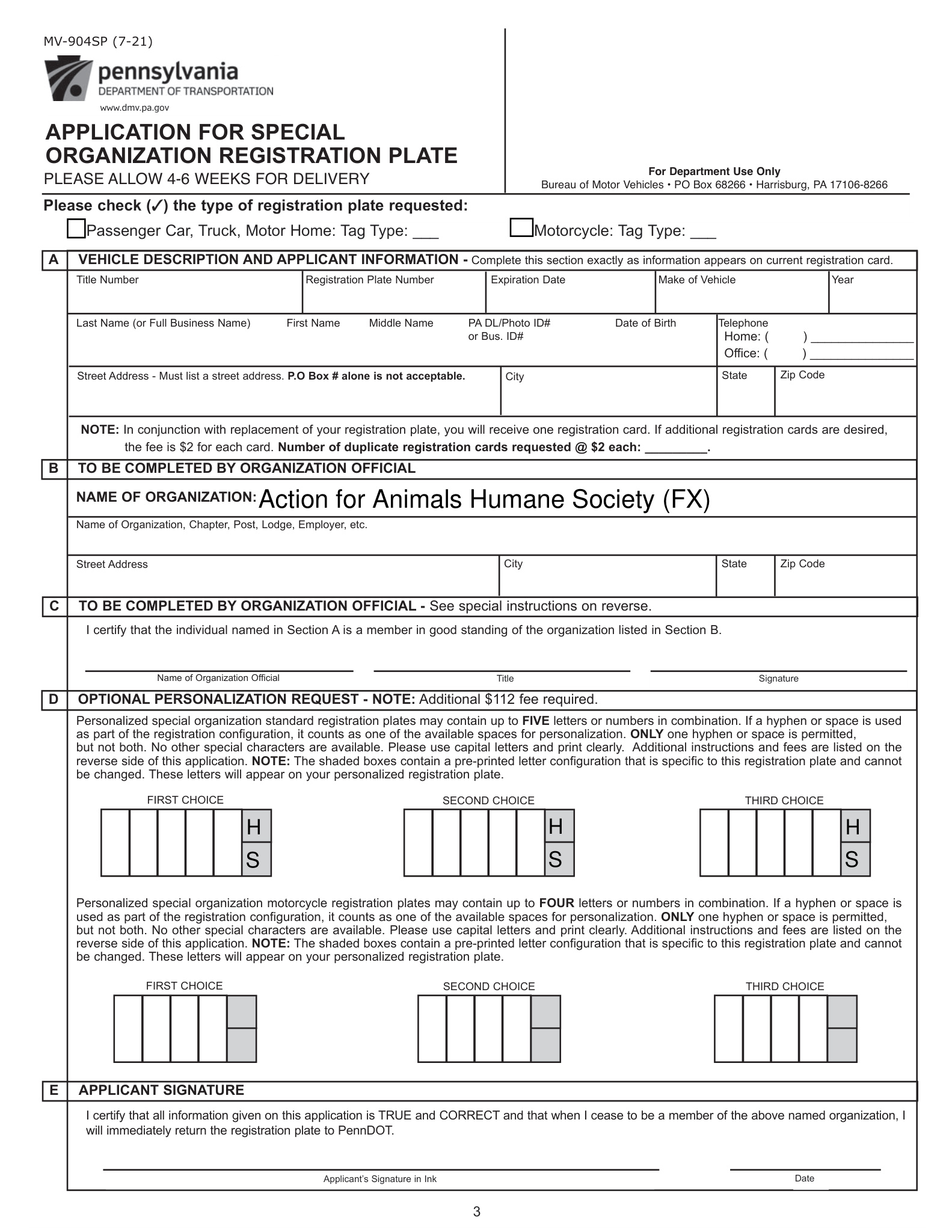} &
\pageimg{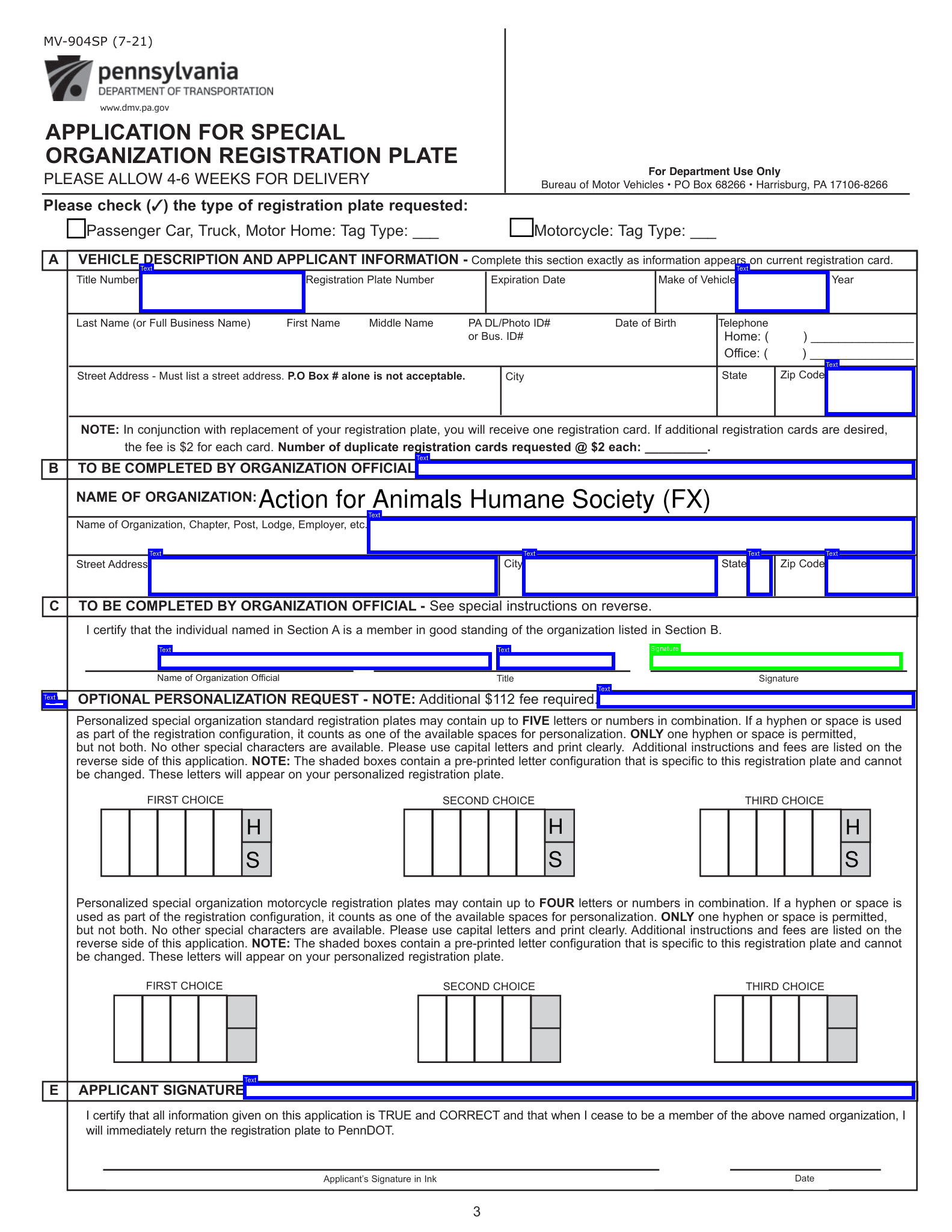} &
\pageimg{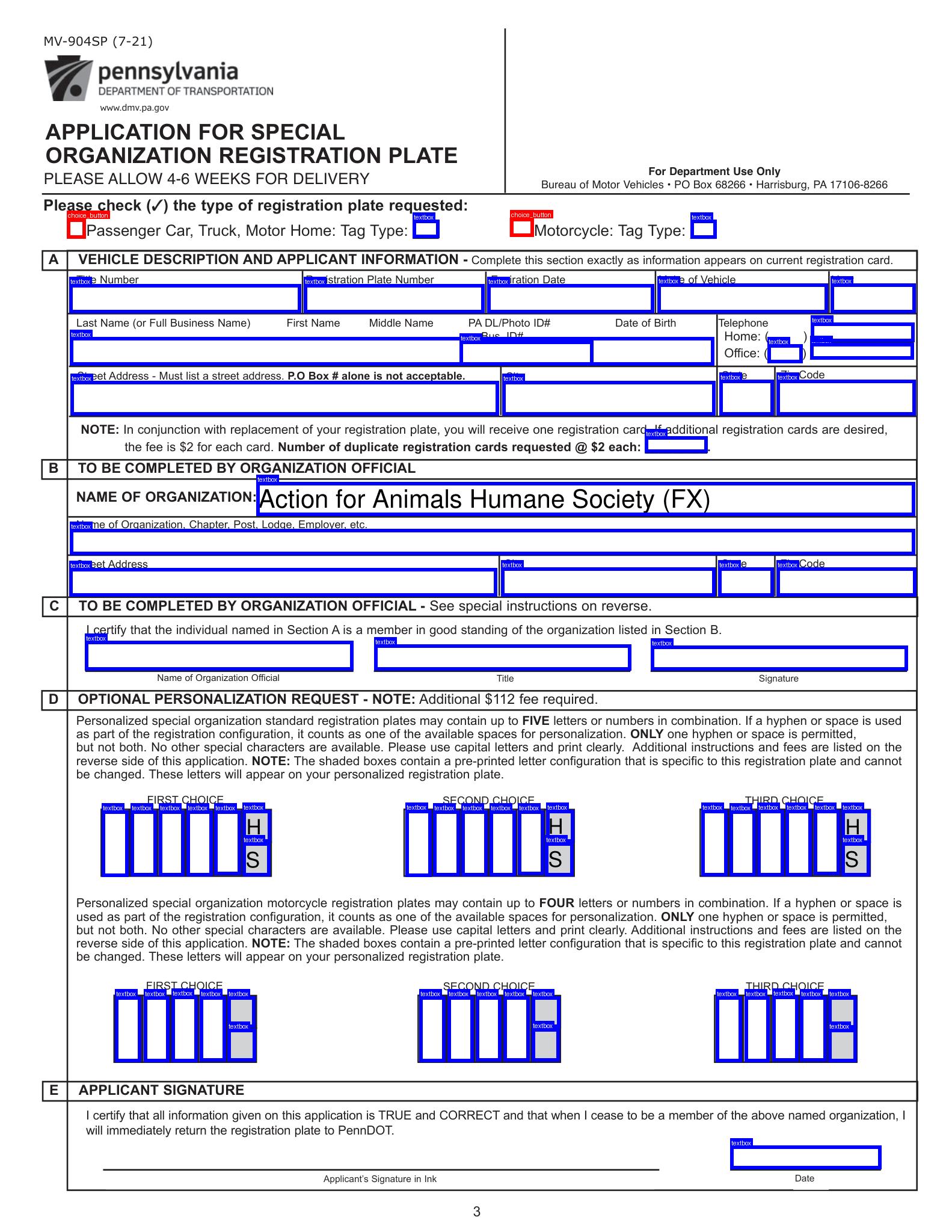} &
\pageimg{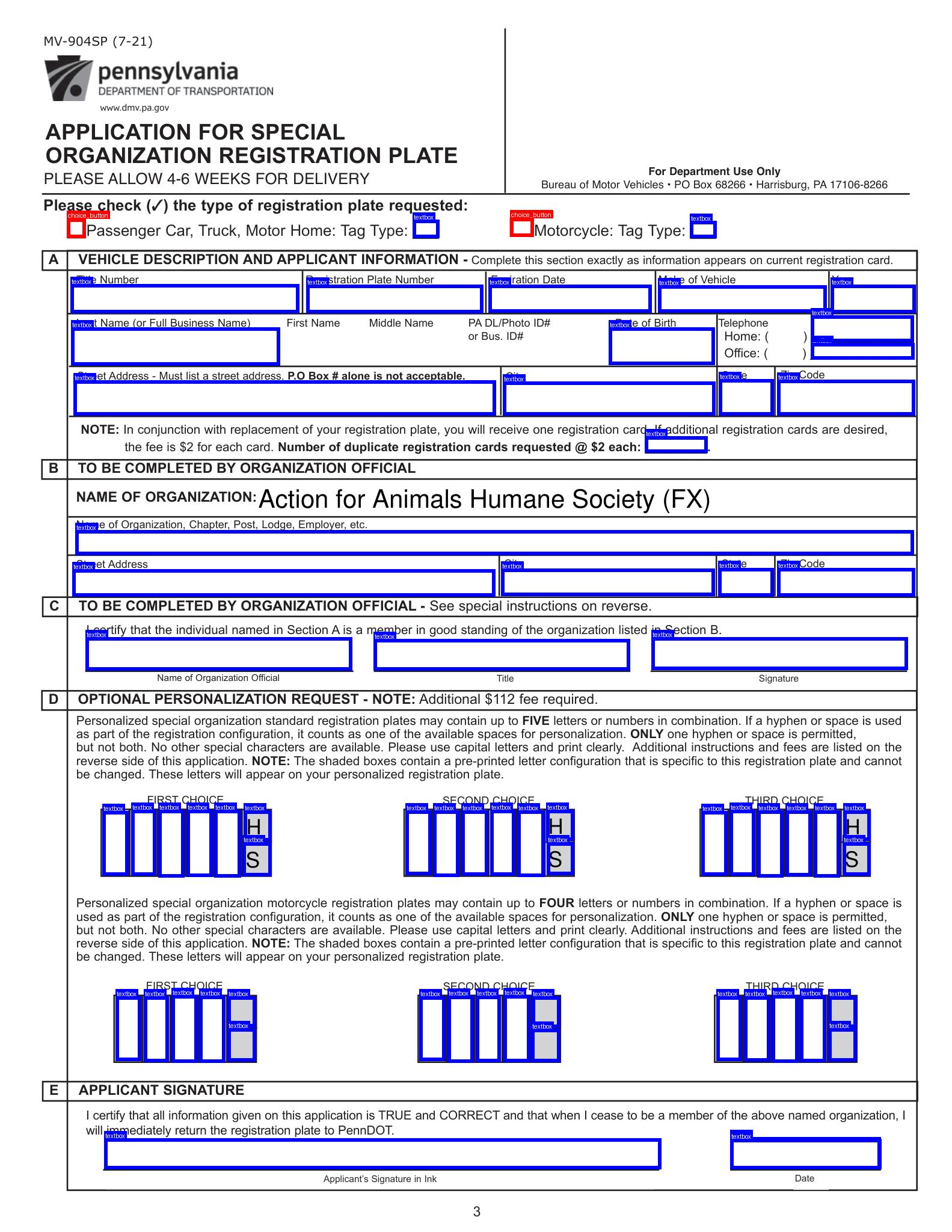} \\

\pageimg{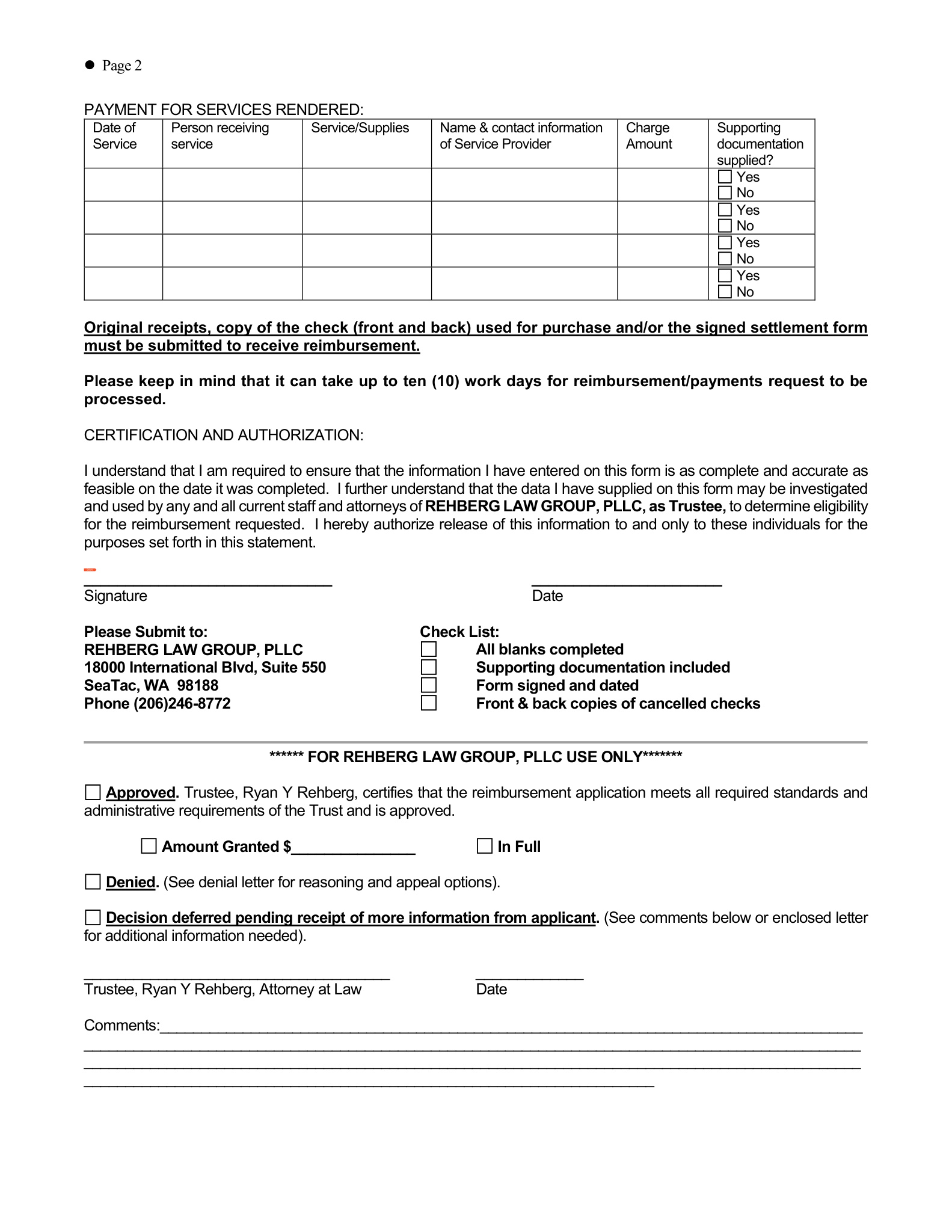} &
\pageimg{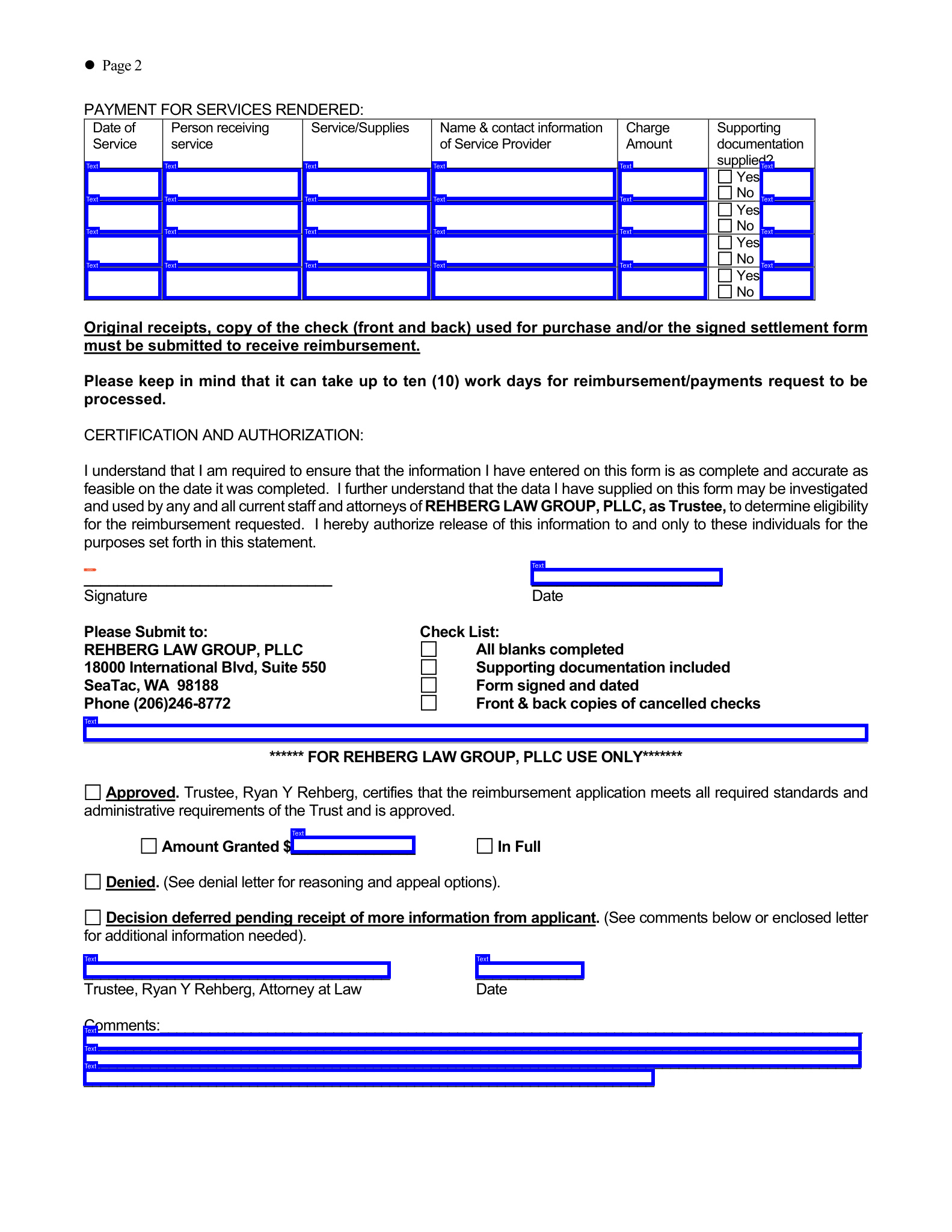} &
\pageimg{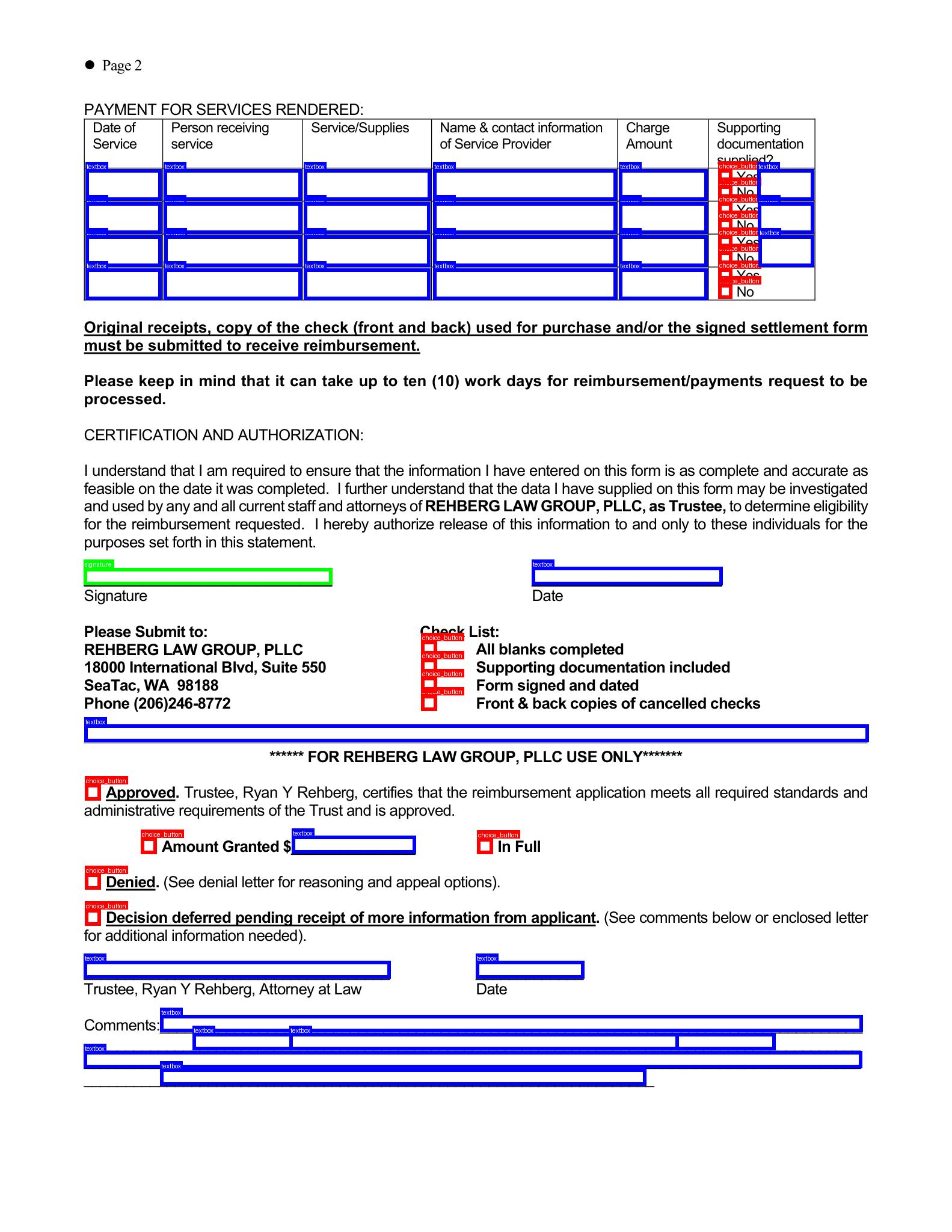} &
\pageimg{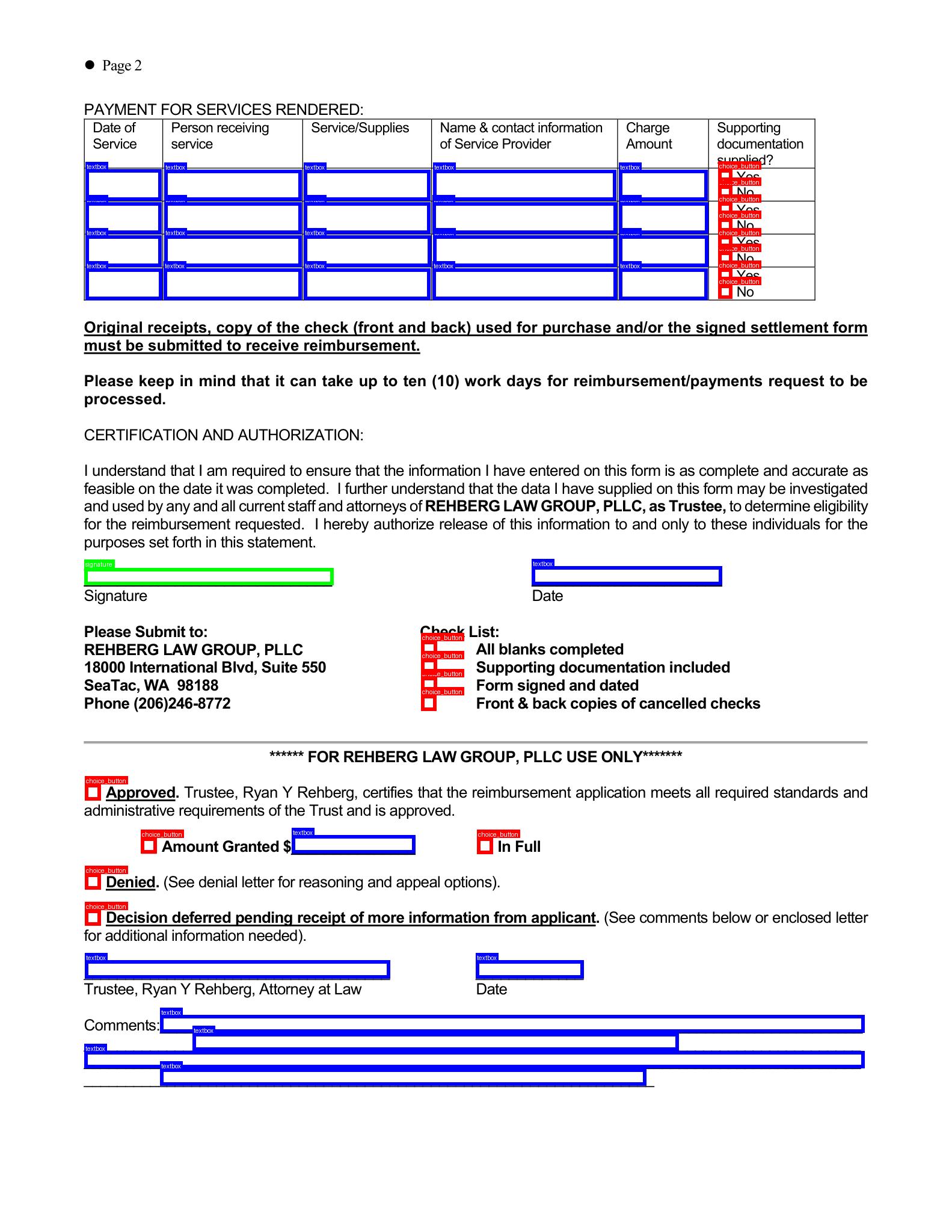} \\

\pageimg{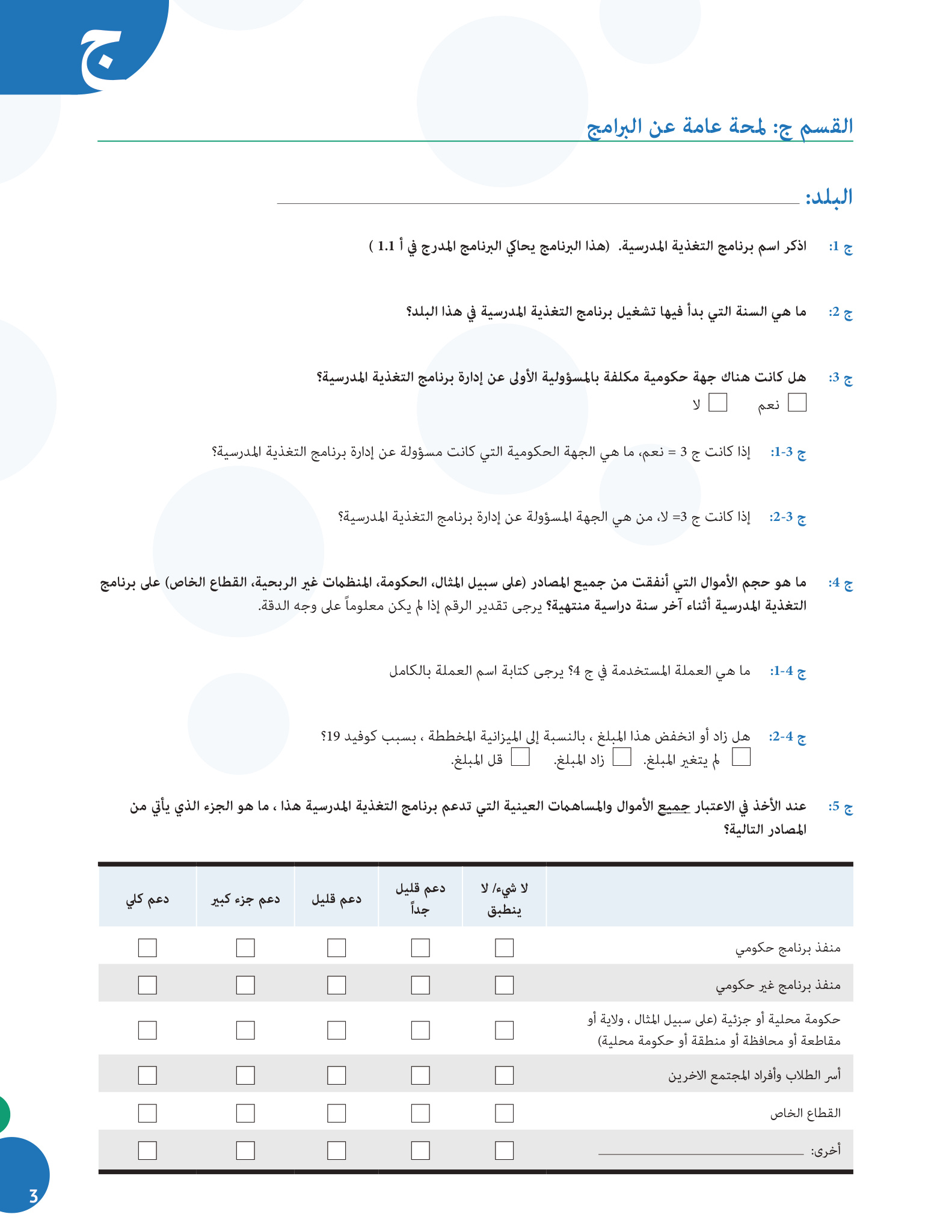} &
\pageimg{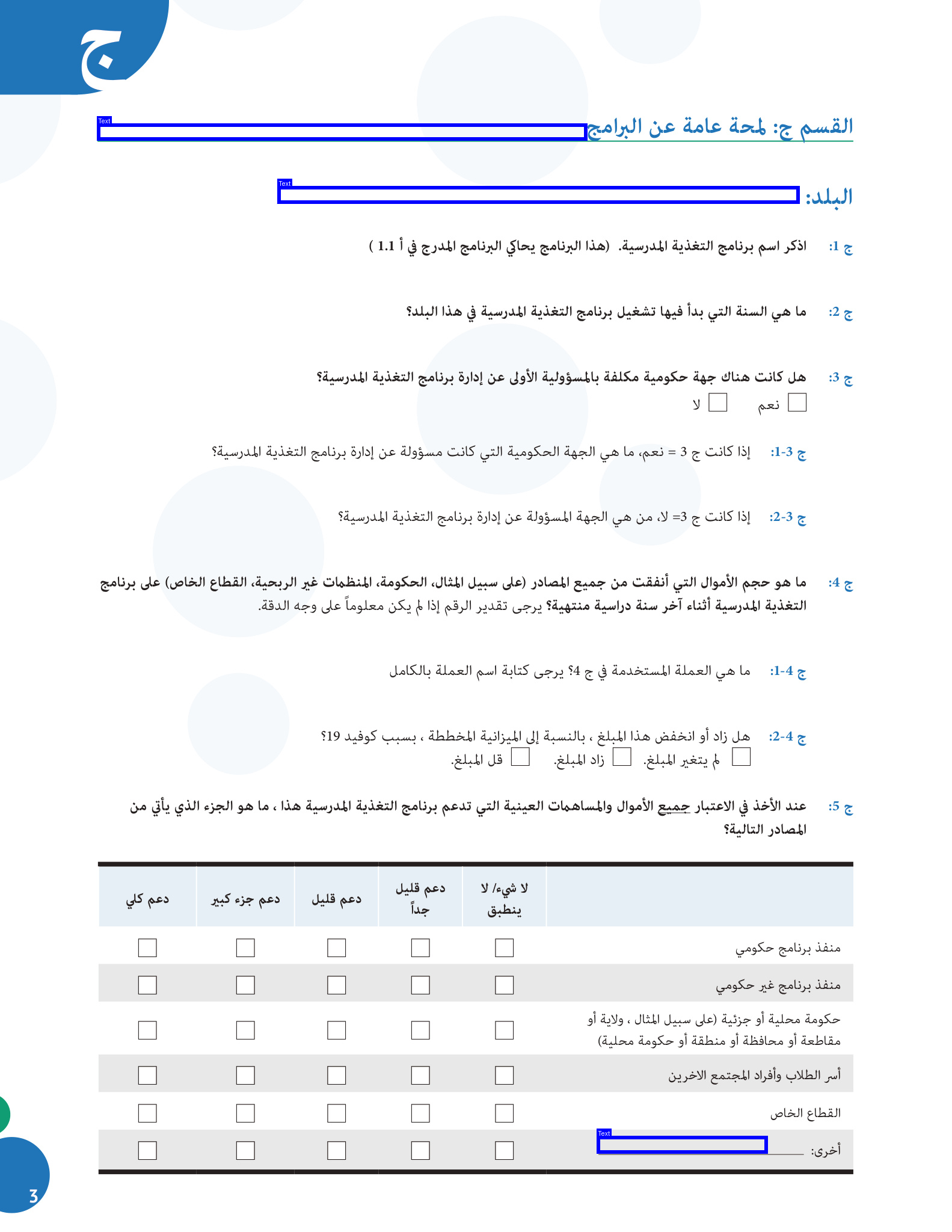} &
\pageimg{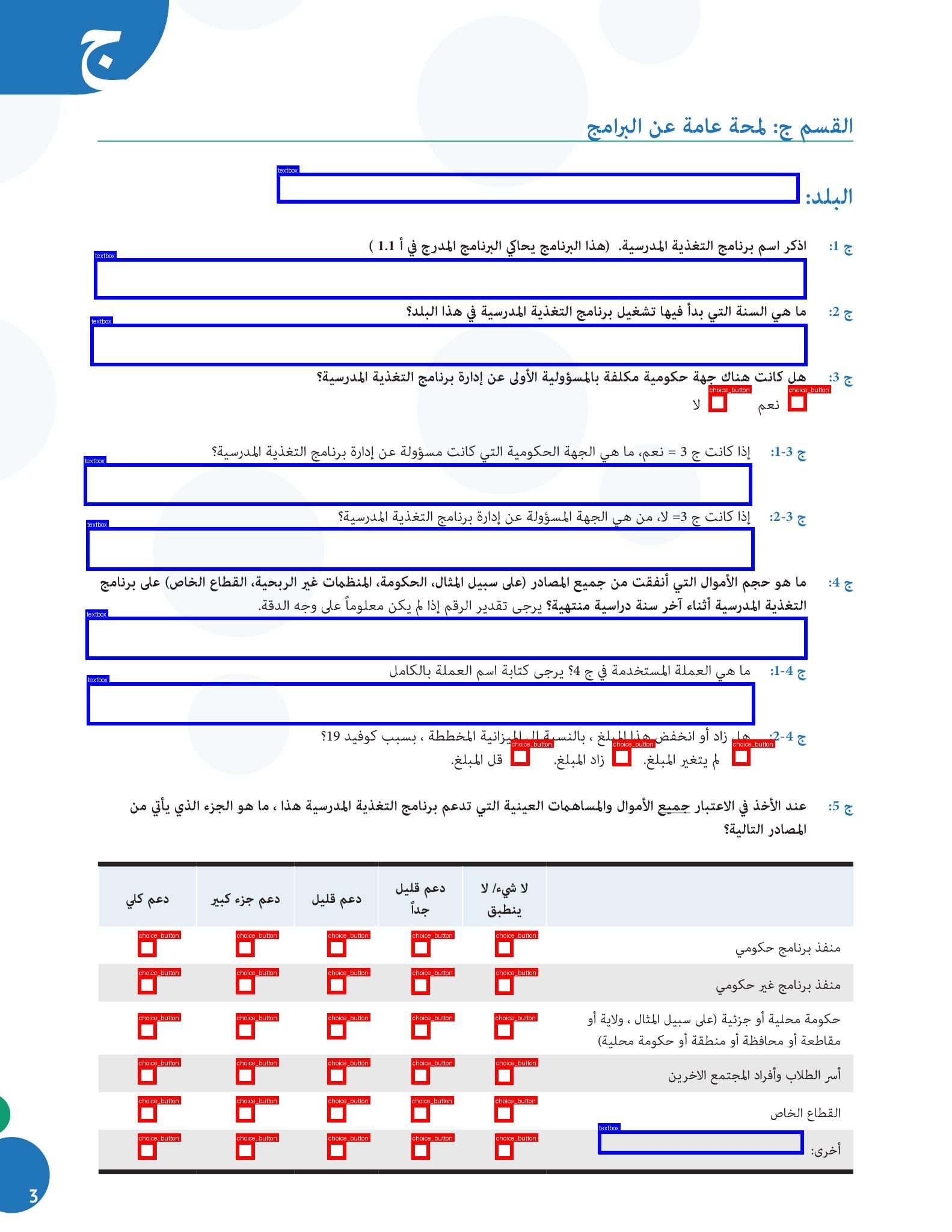} &
\pageimg{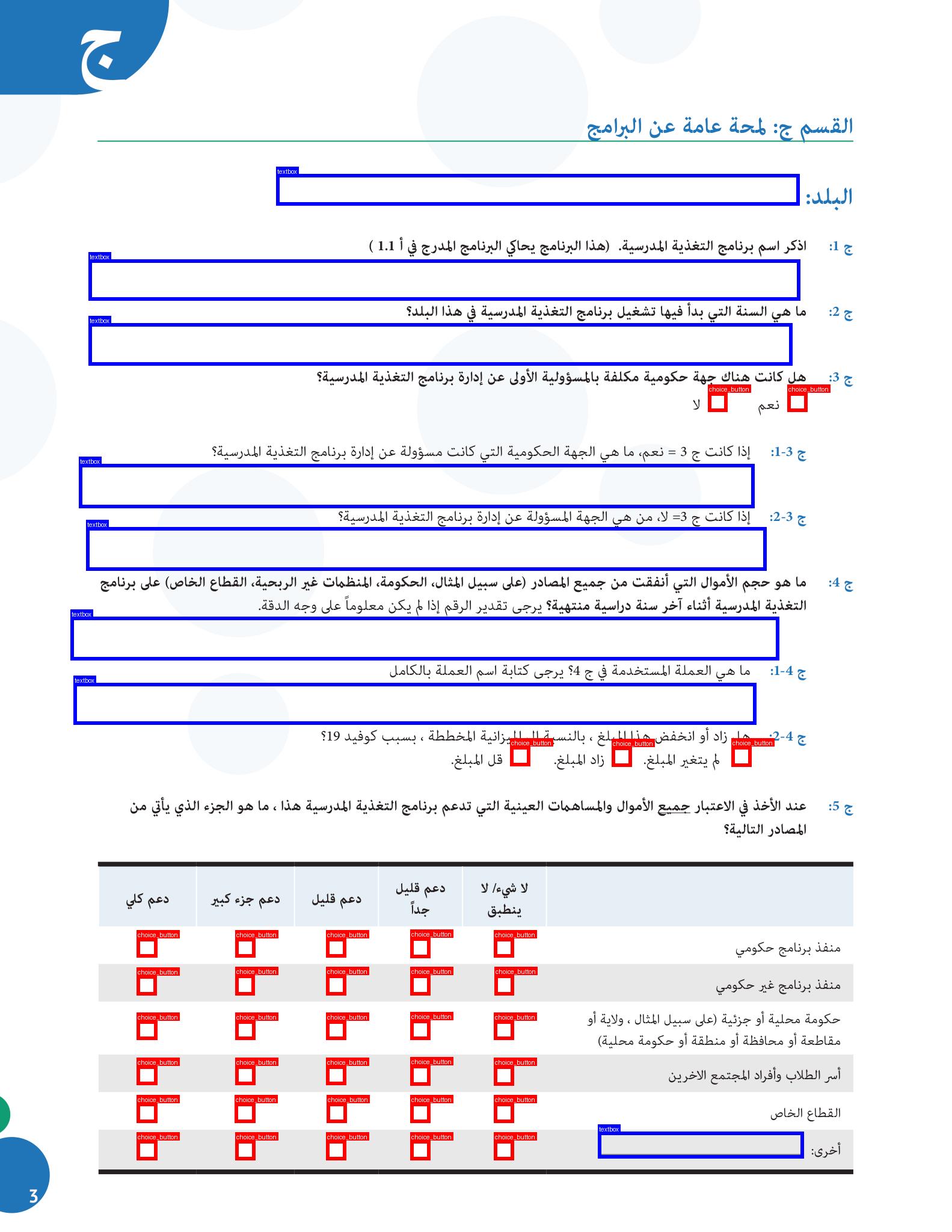} \\

\pageimg{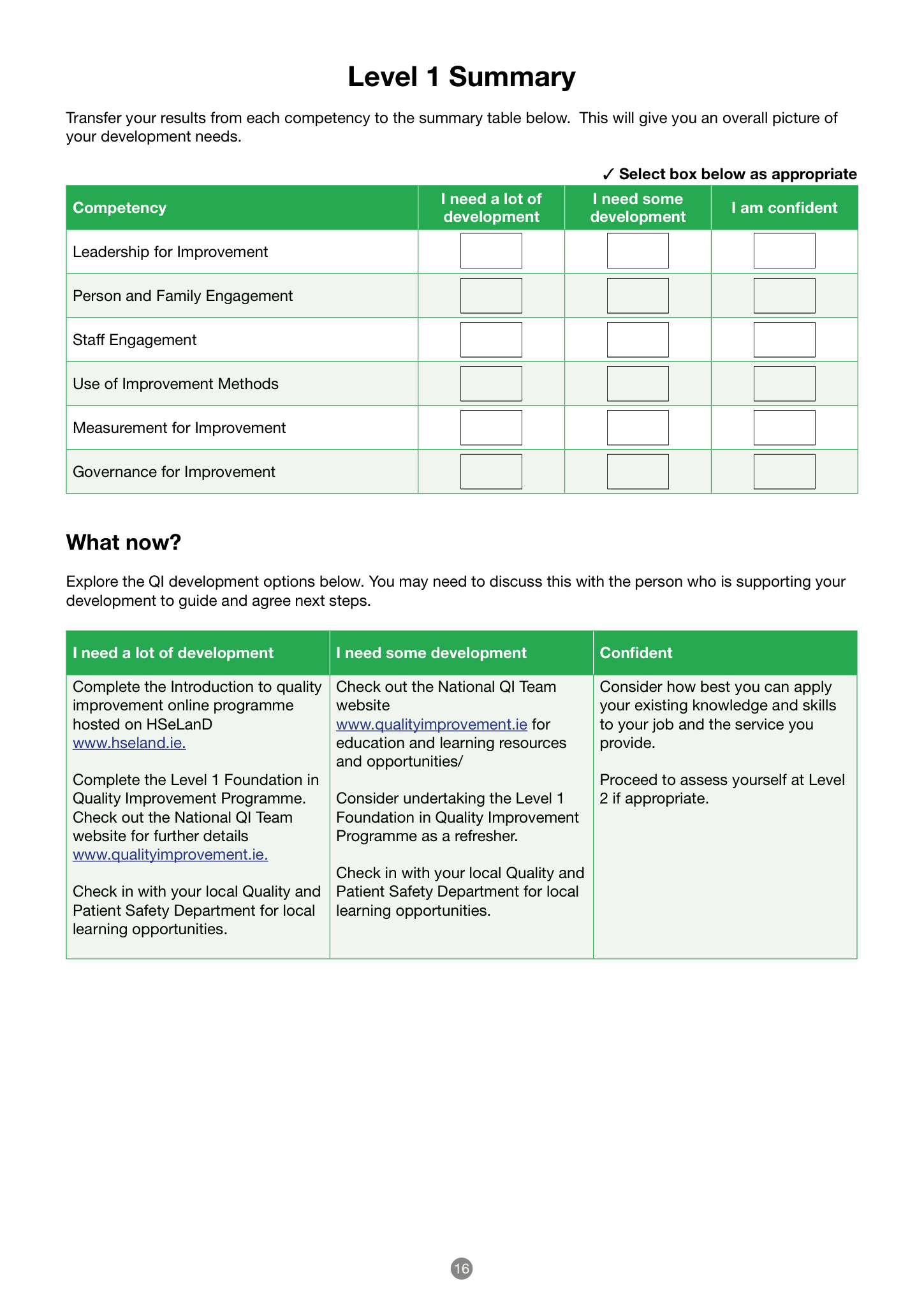} &
\pageimg{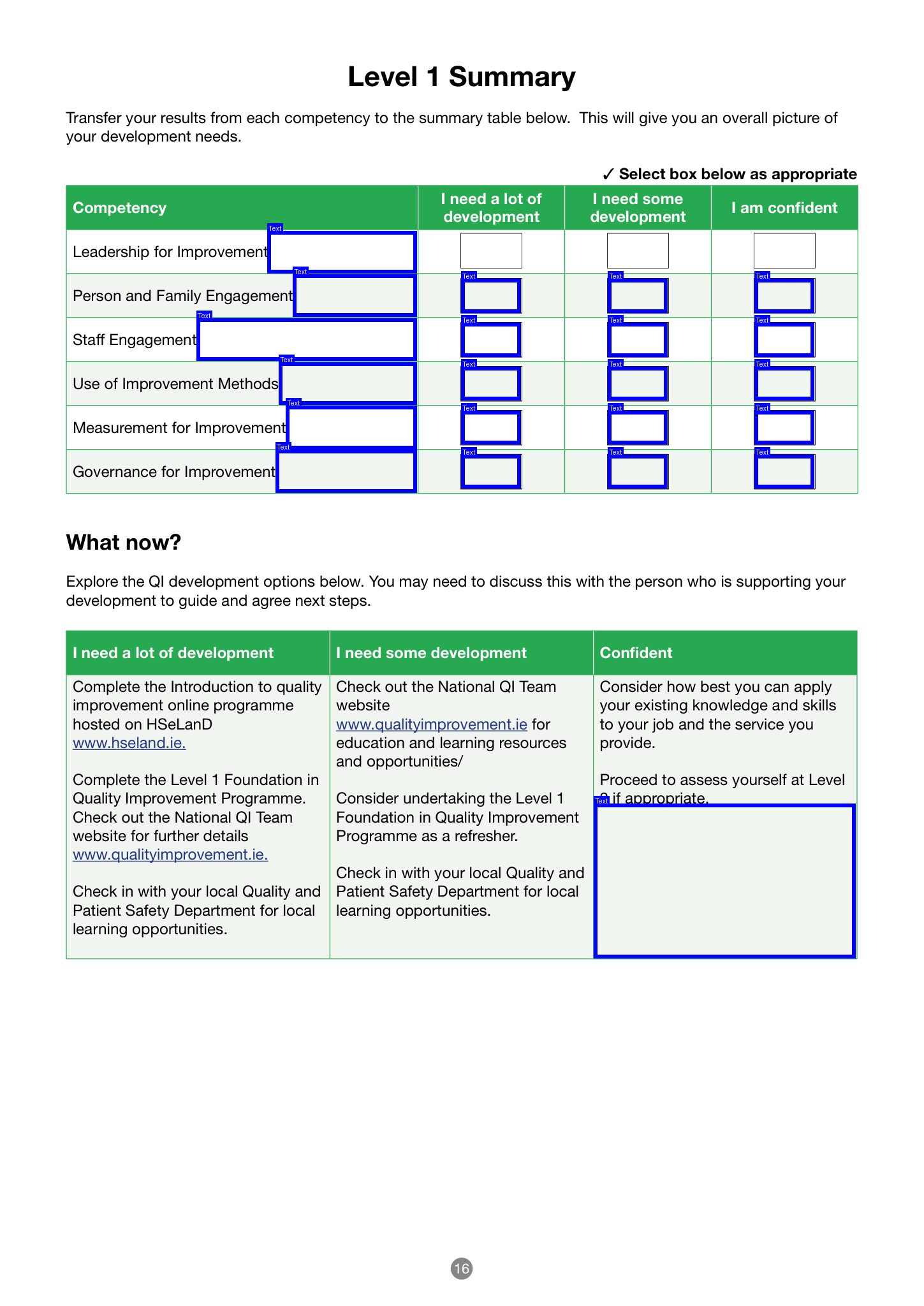} &
\pageimg{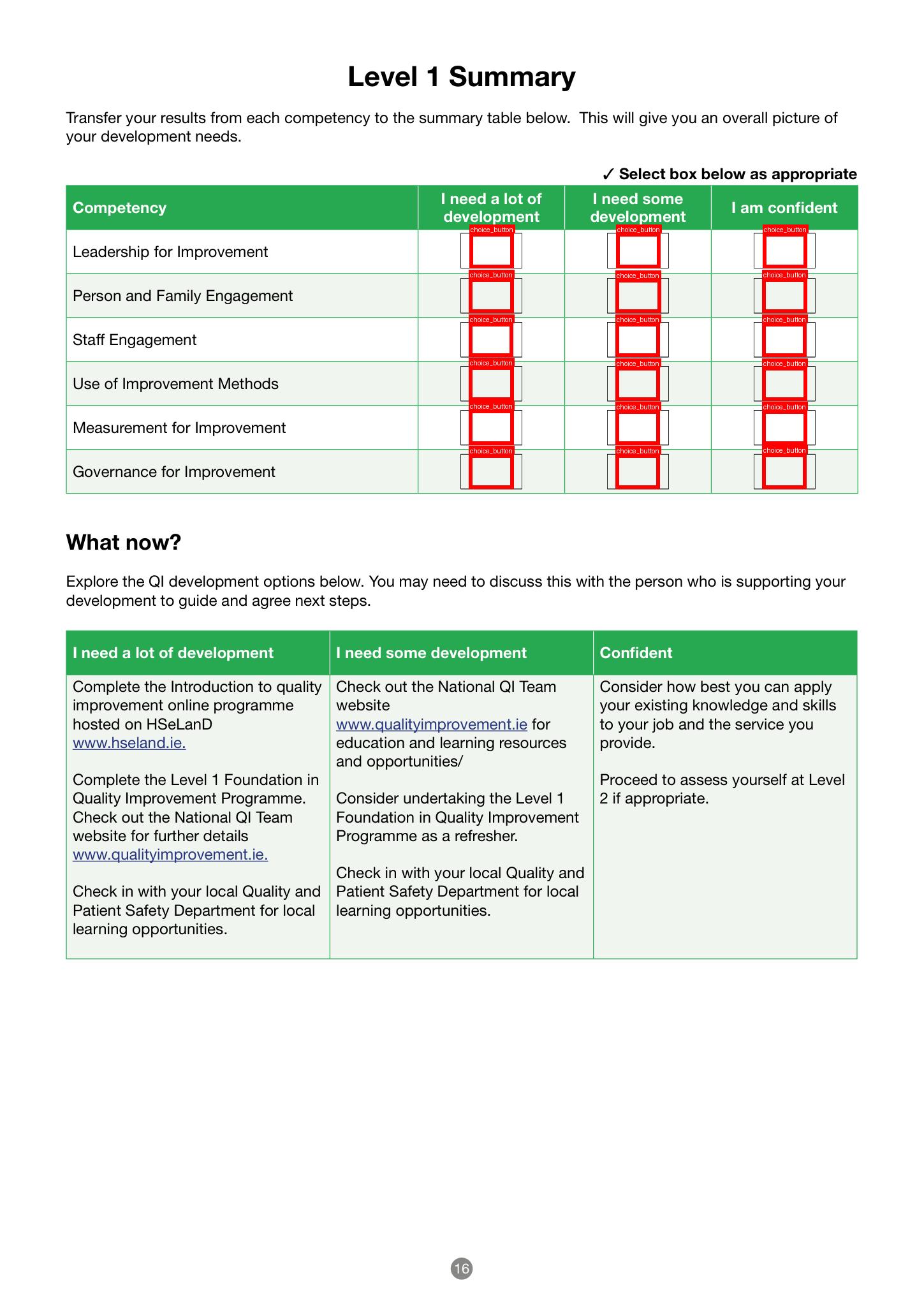} &
\pageimg{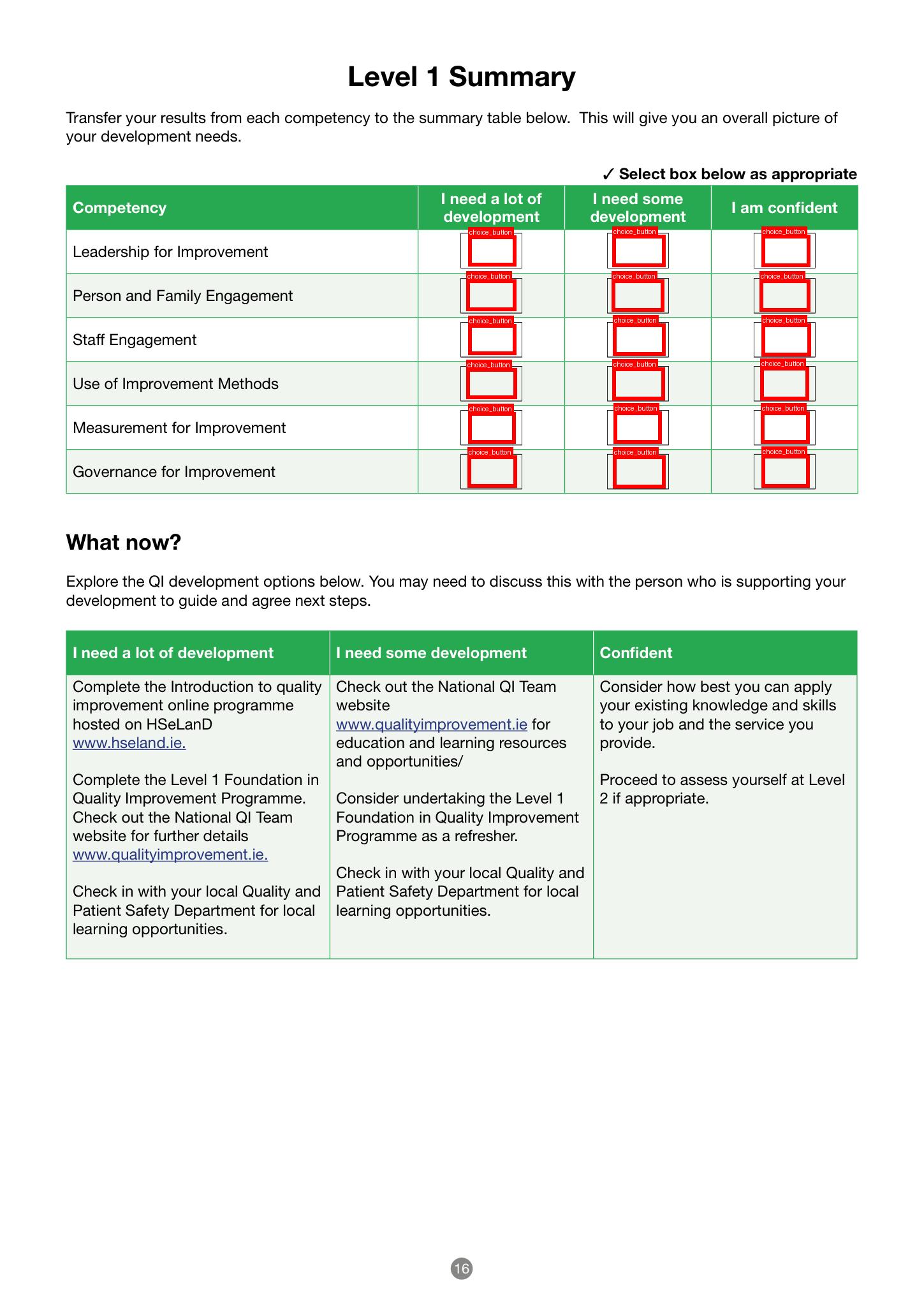} \\
\bottomrule
\end{tabularx}
\end{table*}

\section{Experiments and Results}
\label{sec:results}

We report the performance of FFDNet-S and FFDNet-L in Table~\ref{tab:ffd-results}.
All results are reported as $mAP_{50-95}$.
FFDNet-L consistently outperforms FFDNet-S across all classes.

However, this performance comes at a cost of memory and computatation.
On a single 3090Ti, the inference of FFDNet-L takes roughly 16ms per page, while the inference of FFDNet-S takes roughly 5ms.
FFDNet-S is better suited for mobile or on-device applications, where memory and compute are at a premium.

\begin{table}[t]
  \centering
  \caption{Object detection performance (AP; higher is better) by widget type.}
  \label{tab:ffd-results}
  \setlength{\tabcolsep}{6pt}
  \begin{tabularx}{\columnwidth}{lYYY|Y}
    \toprule
    \textbf{Model} &\textbf{Text} & \textbf{Choice} & \textbf{Sig.} & \textbf{All} \\
    & AP ($\uparrow$) & AP ($\uparrow$) & AP ($\uparrow$) & AP ($\uparrow$) \\
    \midrule
    FFDNet-S (1216px) & 61.5 & 71.3 & 84.2 & 72.3 \\
    FFDNet-L (1216px) & 71.4 & 78.1 & 93.5 & 81.0 \\    
    \bottomrule
  \end{tabularx}
\end{table}

\begin{table}[t]
  \centering
  \caption{Object detection performance}
  \label{tab:resolution}
  \setlength{\tabcolsep}{4pt}
  \begin{tabularx}{\columnwidth}{lYYY|Y}
    \toprule
    \textbf{Resolution} &\textbf{Text} & \textbf{Choice} & \textbf{Sig.} & \textbf{All} \\
    & AP ($\uparrow$) & AP ($\uparrow$) & AP ($\uparrow$) & AP ($\uparrow$) \\
    \midrule
    640px  & 49.2 & 52.2 & 26.7 & 42.7 \\
    960px  & 52.3 & 62.0 & 44.0 & 52.8\\
    1216px & 53.0 & 65.8 & 54.9 & 57.9 \\
    1536px & 53.2 & 67.9 & 65.3 & 62.1 \\
    \bottomrule
  \end{tabularx}
\end{table}

\begin{table}[t]
\centering
\small
\caption{Subcategory results.}
\label{tab:subcategory-ffnet}
\setlength{\tabcolsep}{6pt}
\renewcommand{\arraystretch}{1.15}
\begin{tabularx}{\columnwidth}{X c >{\centering\arraybackslash}p{0.15\columnwidth} >{\centering\arraybackslash}p{0.15\columnwidth}}
\toprule
\textbf{Subcategory} & \textbf{\% Pages} & \textbf{FFNet-S} & \textbf{FFNet-L} \\
 & & AP ($\uparrow$) & AP ($\uparrow$) \\
\midrule
\textbf{All} & 100 & 72.3 & 81.0 \\ \midrule
\multicolumn{3}{l}{\textbf{Language}} \\
\hspace{5mm}English    & 63.6 & 72.4 & 80.6 \\
\hspace{5mm}Cantonese  & 12.6 & 73.4 & 80.4 \\
\hspace{5mm}German     & 6.8 & 68.6 & 80.7 \\
\hspace{5mm}Korean     & 2.6 & 75.6 & 89.3 \\
\hspace{5mm}Spanish    & 2.6 & 62.5 & 78.8 \\
\hspace{5mm}French     & 2.2 & 68.2 & 77.2 \\
\hspace{5mm}Russian    & 1.0 & 33.2 & 69.2 \\
\hspace{5mm}Italian    & 0.9 & 76.8 & 86.1 \\
\hspace{5mm}Portuguese & 0.8 & 75.8 & 84.8 \\
\hspace{5mm}Occitan    & 0.7 & 66.5 & 73.8 \\ \midrule
\addlinespace[2pt]
\multicolumn{3}{l}{\textbf{Domain}} \\ 
\hspace{5mm}Other  & 22.1 & 61.3 & 75.5 \\
\hspace{5mm}Gov't. \& Admin.  & 17.3 & 75.3 & 82.8 \\
\hspace{5mm}Commerce \& Tax       & 11.0 & 70.6 & 78.8 \\
\hspace{5mm}Engineering     & 7.1 & 66.5 & 78.4 \\
\hspace{5mm}Data \& Privacy & 6.1 & 74.3 & 83.0 \\
\hspace{5mm}Law \& Justice  & 6.0 & 72.1 & 79.7 \\
\hspace{5mm}Health  & 5.8 & 70.9 & 77.6 \\
\hspace{5mm}Education & 5.7 & 76.1 & 80.3 \\
\hspace{5mm}Environment  & 5.1 & 80.8 & 84.8 \\
\hspace{5mm}Transportation  & 4.0 & 64.6 & 78.7 \\
\hspace{5mm}Culture \& Religion  & 3.7 & 80.8 & 83.7 \\
\hspace{5mm}Real Estate  & 2.4 & 83.3 & 88.9 \\
\hspace{5mm}Technology  & 2.6 & 75.4 & 79.2 \\
\hspace{5mm}Sports \& Rec.  & 1.1 & 72.0 & 85.7 \\
\bottomrule
\end{tabularx}
\end{table}

\subsection{Resolution Matters In Form Field Detection}
Compared with objects in traditional object detection, many form fields are comparatively fine in an average form.
Certain features that signify where a form element should go, such as underlines or colons, are also very fine.
A consequence of this is that form field detection is more sensitive to input resolution than traditional object detection.

We examine this by finetuning a series of 6 million parameter FFDNet model on a dataset of 10k pages from \commonforms.
We compare 4 resolutions: 640px, 960px, 1216px, and 1536px, and show the results in Table~\ref{tab:resolution}.
Results are reported on the \commonforms test set, so are directly comparable with all other results reported in the paper.

The results show that resolution is tremendously important.
Continuing to 1536px the small models improve in performance across all categories.
The differences in performance are stark; there is roughly a 20 point difference from 640px to 1536px. The \texttt{Choice Button} and \texttt{Signature} form fields are most impacted by resolution.
This makes intuitive sense for \texttt{Choice Button}s, which are often very small objects on a form page (radio buttons or checkboxes).
Distinguishing a signature from a textbox requires a weak form of optical character recognition (OCR) to determine the proximity of indicator words such as ``Signature`` or ``Unterschrift.'' 

The FFDNet models are trained at 1216px, a large size for traditional object detection, but empirically a good trade-off between speed and accuracy for form field detection.

\subsection{FFDNet Outperforms Adobe Acrobat at All Sizes}
We qualitatively compare FFDNet and Adobe Acrobat, with results shown in Table~\ref{tab:qualitative-comparison}.
Of note, Acrobat does not detect choice buttons at all.
Apple Preview also does not detect choice buttons, instead using text inputs in place of all choice buttons.
Acrobat suffers from both low recall, missing tens of form fields per form page, and low precision, table elements and separator lines for text fields.

\subsection{FFDNet is Robust Across Languages/Domains}
Building on the language and domain analysis from Section~\ref{sec:dataset}, we can compare the performance per-language and per-domain.

Both sizes of FFDNet have similar performance across 9 of the 10 most common languages, though they suffer from degraded performance in Russian.
FFDNet-S has a higher variance across languages, as the network is likely not equipped with enough parameters to faithfully capture textual signals across languages.
Similarly, both models perform conssitently across domains, with FFDNet-S having a slightly higher variance than FFDNet-L.

\subsection{Filtering Improves Data Efficiency}
\commonforms employs an aggressive filtering strategy to maximize the likelihood that pages in the dataset are well-prepared forms. 
However, this is done at the cost of potentially more usable data ending up in the training set.
To evaluate this trade-off, we train a 6 million parameter FFDNet model on 10k form pages drawn from the filtered set of forms (59k documents), and the same model on 10k form pages drawn from the set of all forms (760k documents).
The $mAP_{50-95}$ when measured on the test set is roughly 4 points higher (57.9 v. 53.6).
\section{Conclusions and Future Work}
In this paper we build and release \commonforms, a dataset of 490k diverse form images filtered from PDFs in Common Crawl.
We show that the filtering process improves the data efficiency versus the strategy of keeping all forms.
Using this dataset, we train a family of high-resolution object detectors, FFDNet-S and FFDNet-L.
We show that FFDNet-L qualitatively yet clearly outperforms Adobe Acrobat at form field detection on a test set of forms.
We release the preparation code, dataset, and models open source.

In future work, we seek to tackle the complete problem of form preparation, by building datasets and models that capture the semantics of forms.
In addition, we believe that there are several potential avenues to improve FFDNet, including bringing in recent work in object detection.
In particular, performance on scans and foreign language documents can be improved, possibly via some form of data augmentation or resampling. 
There are likely to be gains cleaning up the form preparation inconsistencies noted in Section~\ref{sec:inconsistencies}.
{
    \small
    \bibliographystyle{ieeenat_fullname}
    \bibliography{main}
}

\end{document}